\documentclass[10pt,twocolumn,letterpaper]{article}

\usepackage[pagenumbers]{cvpr}              % To produce the CAMERA-READY version
\usepackage{graphicx}
\usepackage{amsmath}
\usepackage{amssymb}
\usepackage{booktabs}

\usepackage[pagebackref,breaklinks,colorlinks]{hyperref}
\usepackage[accsupp]{axessibility}
\usepackage[ruled]{algorithm2e}
\usepackage{bbm}
\usepackage{latexsym}

\usepackage{epsfig}
\usepackage{graphicx}
\usepackage{amsmath,bm}
\usepackage{amssymb}
\usepackage{bbding}
\usepackage{color}
\usepackage{xcolor}
\usepackage{booktabs}
\usepackage{etoolbox} 
\usepackage{etoolbox} 
\newcommand{\MyMapTemplatePrefixc}[4]{\expandafter#1\csname#3#4\endcsname{#2{#4}}} % it remembles a template: \#3#4 --> #2{#4}
\forcsvlist{\MyMapTemplatePrefixc {\def} {\mathcal}{c}} {A,B,C,D,E,F,G,H,I,J,K,L,M,N,O,P,Q,R,S,T,U,V,W,X,Y,Z}  % e.g., \cA --> \mathcal{A}

\newcommand{\MyMapTemplatePrefixtb}[5]{\expandafter#1\csname#4#5\endcsname{#2{#3{#5}}}} % it remembles a template: \#3#4 --> #2{#4}
\forcsvlist{\MyMapTemplatePrefixtb {\def} {\tilde}{\mathbf}{t}} {A,B,C,D,E,F,G,H,I,J,K,L,M,N,O,P,Q,R,S,T,U,V,W,X,Y,Z}  % e.g., \tA --> \tilde{A}
\forcsvlist{\MyMapTemplatePrefixtb {\def} {\tilde}{\mathbf}{t}} {0,1,a,b,c,d,e,f,g,h,i,j,k,l,m,n,o,p,q,r,s,u,v,w,x,y,z}  % e.g., \ta --> \tilde{a}

\newcommand{\MyMapTemplateNoPrefix}[3]{\expandafter#1\csname#3\endcsname{#2{#3}}}
\forcsvlist{\MyMapTemplateNoPrefix {\def} {\mathbf} } {0,1,a,b,c,d,e, f, g, h, i, j, k, l, m, n, o, p, q, r, u, v, w, x, y, z} % e.g., \a --> \mathbf{a}
\forcsvlist{\MyMapTemplateNoPrefix {\def} {\mathbf} } {A,B,C,D,E,F,G,H,I,J,K,L,M,N,O,P,Q,R,S,T,U,V,W,X,Y,Z}  % e.g., \A --> \mathcal{A}

\def\ie{\emph{i.e.}}
\def\eg{\emph{e.g.}}

\def\wrt{\emph{w.r.t.}}

\usepackage[capitalize]{cleveref}
\crefname{section}{Sec.}{Secs.}
\Crefname{section}{Section}{Sections}
\Crefname{table}{Table}{Tables}
\crefname{table}{Tab.}{Tabs.}

\def\cvprPaperID{2645} % *** Enter the CVPR Paper ID here
\def\confName{CVPR}
\def\confYear{2022}

\begin{document}

\title{Revisiting AP Loss for Dense Object Detection: Adaptive Ranking Pair Selection}

\author{Dongli Xu, Jinhong Deng, Wen Li \thanks{\ Corresponding author.}\\
School of Computer Science and Engineering \& Shenzhen Institute for Advanced Study\\
University of Electronic Science and Technology of China\\
{\tt\small \{dongliixu,jhdeng1997,liwenbnu\}@gmail.com}
% \and
}
\maketitle

%%%%%%%%% ABSTRACT
\begin{abstract}
Average precision (AP) loss has recently shown promising performance on the dense object detection task. However, a deep understanding of how AP loss affects the detector from a pairwise ranking perspective has not yet been developed.
In this work, we revisit the average precision (AP) loss and reveal that the crucial element is that of selecting the ranking pairs between positive and negative samples. Based on this observation, we propose two strategies to improve the AP loss.
The first of these is a novel Adaptive Pairwise Error (APE) loss that focusing on ranking pairs in both positive and negative samples. Moreover, we select more accurate ranking pairs by exploiting the normalized ranking scores and localization scores with a clustering algorithm.
Experiments conducted on the MS-COCO dataset support our analysis and demonstrate the superiority of our proposed method compared with current classification and ranking loss.
The code is available at https://github.com/Xudangliatiger/APE-Loss.
\end{abstract}

%%%%%%%%% BODY TEXT
\section{Introduction}
\label{sec:intro}

\begin{figure*}[ht]
\vspace{-0.1cm}
\setlength{\abovecaptionskip}{0cm}
\setlength{\belowcaptionskip}{0cm}
    \centering
    \includegraphics[width=0.9\linewidth]{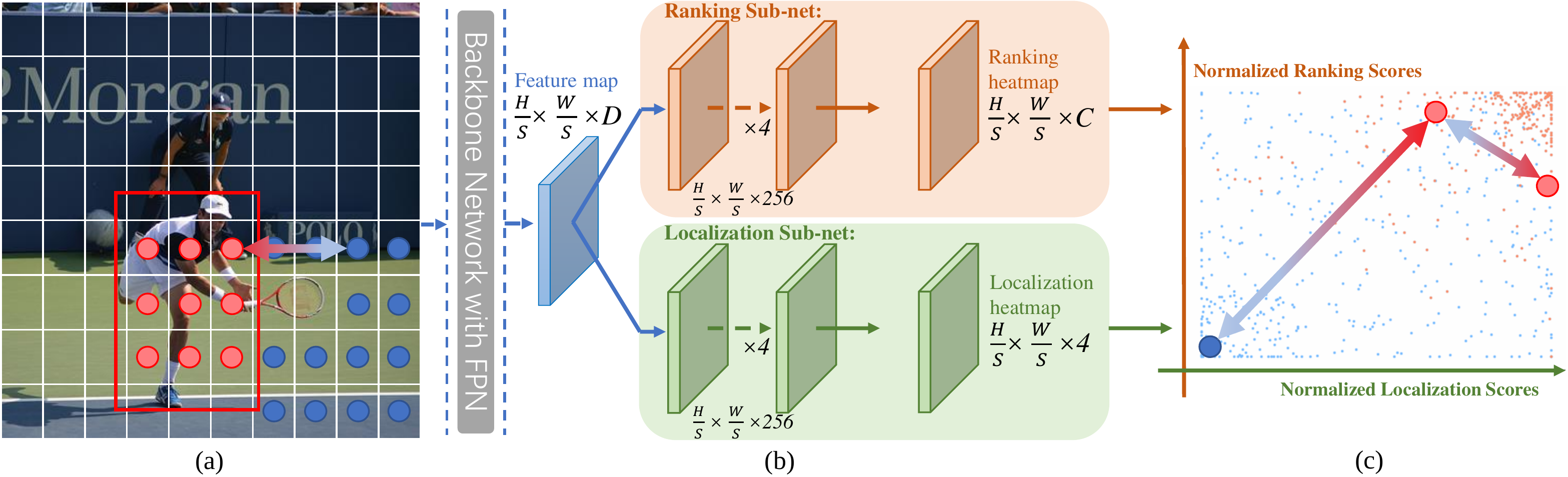}
    \caption{Comparison between our ranking pair selection and traditional methods. Here, the large red and blue dots are used to highlight positive and negative samples, respectively, while the two-way arrows are used to highlight typical ranking pairs. (a) Current pairwise ranking methods ignore the pair between positive positions and focus on the anchor-based IoU to select ranking pairs. The red and white boxes are the ground truth and preset anchor. (b) FCOS detector (this paper omits the center-ness branch). (c) Our method can make up for the loss of ranking pairs between positive samples and focus on the distribution of prediction scores to adaptively select more accurate ranking pairs.}
\vspace{-0.3cm}
\end{figure*}

Object detection is one of the fundamental computer vision tasks and aims to predict both the category labels and the bounding-box coordinates of all objects in a given image~\cite{ren2015fasterrcnn,liu2016ssd,Lin_2017_retina,cai2018cascade,tian2019fcos}.
It also plays an important role in many down-stream applications such as instance segmentation~\cite{he2017mask,bolya2019yolact,xie2020polarmask,wang2020solo} and face detection~\cite{Deng_2020retinaface}. 
Modern object detectors can be divided into two-stage approaches~\cite{ren2015fasterrcnn,cai2018cascade} and one-stage approaches~\cite{xiangyu2018shuffle,yxiong2021mobiledets}.
One-stage detectors adopt dense prediction without a region proposal phase; thus they are also known as dense object detectors.
One-stage object detectors are also naturally faster than two-stage detectors and are popular in real-world applications such as that on edge devices~\cite{xiangyu2018shuffle,yxiong2021mobiledets}.
Most one-stage object detectors~\cite{liu2016ssd,redmon2016yolo,Lin_2017_retina,tian2019fcos,zhu2019fsaf,zhang2020atss,li2020generalized} rely on a classification task to discriminate the category of objects, \ie, directly predicting the category probability for every patch in an image.
However, it often suffers from an extreme imbalance problem, as huge numbers of background patches~(\ie, negative samples) can overwhelm the sufficient loss gradients for foreground patches~(\ie, positive samples).
Recently, to alleviate this sensitivity to the ratio of negative and positive samples, Average Precision Loss~\cite{cheng2019aploss} converts the classification task into a ranking task by explicitly modeling sample relationships that are calculated by comparing all sample pairs.

Despite AP loss performing well at addressing the sample imbalance problem, the essential mechanism of how AP loss affects the detector is still veiled.
Accordingly, in this paper, we delve into AP loss from a pairwise ranking perspective~\cite{burges2005ranknet}; specifically, the crucial part of AP loss is the minimizing of the pairwise error between the positive and negative samples. 
To find out the essential properties of AP loss, we analyze components of pairwise error separately and reveal that the key factor is the accurate and complete ranking pair selection.
Consequently, we improve the AP loss by adjusting the way in which ranking pairs are constructed and selected.
Proper ranking pair selection achieves notable performance gains for object detection.

More specifically, we first derive a reformulation of the AP loss and show that it contains three main components of generalized pairwise error: distance function, balance constant and ranking pair selection.
We conduct detailed experiments to verify these different components' effects and determine that proper ranking pair selection plays a vital role in producing accurate detection results.
We then investigate the barriers to good performance encountered by existing ranking pair selection strategies and identify the following issues:
(1) Traditional strategies~\cite{Lin_2017_retina,tian2019fcos,zhang2020atss,kim2020paa} ignore the ranking pairs between positive samples, which may lead to inaccurate category probability prediction, and harms the Non-Maximum Suppression~(NMS) result;
(2) The probability of image content in the ranking task and accuracy in the localization task,~\ie, ranking score and localization score have different distributions, which brings imbalanced attentions to those two tasks during the pair selection processing as in Probabilistic Anchor Assignment (PAA)~\cite{kim2020paa}.

To alleviate these barriers, we propose a simple yet effective Adaptive Ranking Pair Selection~(ARPS) approach to provide complete and more accurate ranking pairs for calculating pairwise error.
To begin with, we build extra ranking pairs from the positive set to calculate Adaptive Pairwise Error~(APE), which is differentiable and easy to implement and can exploit the ranking information between positive samples.
It is worth noting that APE loss can also be considered a more accurate AP loss formulation.
Second, we align the instance-level ranking scores and localization scores via normalization and feed them into a clustering algorithm~(\eg Gaussian Mixture Model in PAA~\cite{kim2020paa}) to create a better split between the positive and the negative set.
After that, ranking pairs can be easily obtained from every combination pair of two clusters.
Experimental results show that our method can optimize the training procedure of ranking tasks with higher accuracy.

The contributions of this work are three-fold and can be summarized as follows:
1) We conduct thorough experiments to verify each part of AP loss from a pairwise perspective and reveal that inappropriate ranking pair selection is the main obstacle;
2) we propose a ranking pair selection algorithm, \ie, ARPS, to exploit complete and more accurate ranking pairs automatically;
3) our method achieves competitive performance when evaluated against all other existing classification and ranking methods.

%-------------------------------------------------------------------------

\section{Related Work}
\noindent \textbf{Dense Object Detection:}
In object detection frameworks, two-stage detectors commonly utilize cross-entropy loss ~\cite{ren2015fasterrcnn,cai2018cascade} for the classification purpose, while many dense detectors equip GHM loss~\cite{li2019ghm} and focal loss~\cite{Lin_2017_retina} to alleviate the sample imbalance problem.
Moreover, a large number of works have attempted to boost dense object detection, using methods that include exploiting information from different layers~\cite{Lin_2017_fpn,Ghiasi_2019_nasfpn}, feature alignment~\cite{H_2020_borderdet,zhang2020vfl} and accurate category probability~\cite{Jiang_2018_iounet,kim2020paa,xu2020asso,li2020generalized,zhang2020vfl}.

\noindent \textbf{Accurate Object Detection:} Recent studies~\cite{Jiang_2018_iounet,tian2019fcos,xu2020asso,li2020generalized} have integrated the quality information of positive samples, \ie, Intersection over Union (IoUs) between predicted boxes and ground-truth boxes, into the classification task or additional IoU~(or center-ness) prediction task~(\eg, IoUs are employed as the labels for the CE loss on classification task).
When detectors are able to predict the qualities of the positive samples, more accurately predicted category probabilities (\eg, IoU-aware classification scores
in~\cite{zhang2020vfl}) will greatly improve the results of the box selection algorithm  NMS~\cite{Jiang_2018_iounet}.

\noindent \textbf{Ranking for Object Detection:}
Recently, AP loss~\cite{cheng2019aploss} has been proposed to avoid the imbalance problem in dense object detection by proposing a novel ranking framework to replace the classification task.
aLRP Loss~\cite{oksuz2020alrp} extends AP loss to optimize unified detection metric LRP and also addresses the localization task. DR loss~\cite{qian2020dr} re-distributes positive and negative samples so that only one expectation ranking pair is used in loss; however, it ignores localization qualities,~\ie, IoUs of positive samples. 
Learning to ranking model~\cite{tan2019l2r} adds an extra branch in two-stage detectors to solve the ranking of localization quality. 
Similarly, RankDetNet~\cite{Liu_2021_rankDetNet} considers the task as a separated ranking task and tackles it with an IoU-guided ranking loss.
Adopting a different approach to these works, we verify each part of AP loss from a pairwise perspective through experimental analysis and determine that inappropriate ranking pair selection is the main obstacle. Hence we propose a ranking pair selection algorithm, ARPS, to automatically produce complete and more accurate ranking pairs and merge the localization qualities into a more accurate AP metric optimization problem.

From the perspective of localization quality ranking, a related and concurrent work is RS loss~\cite{Oksuz_2021_ranksort}, which also proposes to leverage the information among positive samples by applying a sorting loss.
Although RS loss~\cite{Oksuz_2021_ranksort} considers the relation among positive samples, it differs from our proposed method in essential ways.
They combine IoUs into error-driven update to tackle the non-differentiable nature of ranking and sorting; by contrast we utilize IoUs information to construct extra ranking pairs from positive samples into a differentiable pairwise error, and can work without error-driven update.

\noindent \textbf{Training Sample Selection:} Throughout this paper, we will use the terms ``ranking pair selection'' and ``training sample selection'' interchangeably, with the understanding that the ranking pairs are formulated as different pairs from two sample sets; \ie, the positive and negative. RetinaNet~\cite{Lin_2017_retina} defines positive samples using a manually adjusted IoU threshold of preset anchor boxes, while ATSS~\cite{zhang2020atss} provides an adaptive IoU threshold. Recent studies~\cite{li2020noisyanchor,kehuang2020mal,kim2020paa} adaptively separate positive and negative sets depending on the model’s learning status.
%-------------------------------------------------------------------------

\section{Preliminary}

Based on traditional detector designs, let $\mathbf{O} \in R^{H \times W \times 3}$ be an input image of width $W$ and height $H$.
Given an image $\O$, we use a backbone network $\mathcal{F}_{FPN}$ with Feature Pyramid Net~(FPN)~\cite{Lin_2017_fpn} to obtain $n$ feature maps as follows: 
\begin{equation}
\label{eq.1}
\begin{small}
\mathbf{F}_0,\mathbf{F}_1,...,\mathbf{F}_i,...,\mathbf{F}_n = \mathcal{F}_{FPN} (\mathbf{O})
\end{small}
\end{equation}
where $\mathbf{F}_i\in R^{\frac{H}{S_i}\times \frac{W}{S_i} \times D_i}$ is the feature map at layer $i\in \{0,1,...,n\}$, $S_i$ is the output stride and $D_i$ is the dimensional depth of $\mathbf{F}_i$.
The ranking sub-net $\mathcal{F}_{rank}$  and localization sub-net $\mathcal{F}_{loc}$ are applied on each $\mathbf{F}_i$, such that our objective heatmaps are obtained as follows:
\begin{equation}
\label{eq.2}
\begin{small}
\begin{aligned}
\hat{\mathbf{P}}_i = \mathcal{F}_{rank} (\mathbf{F}_i) \qquad \hat{\mathbf{B}}_i = \mathcal{F}_{loc} (\mathbf{F}_i)
\end{aligned}
\end{small}
\end{equation}
where $\hat{\mathbf{P}}_i \in R^{\frac{H}{S_i}\times \frac{W}{S_i} \times M} $ and $\hat{\mathbf{B}}_i \in R^{\frac{H}{S_i}\times \frac{W}{S_i} \times 4} $ represent category probability prediction and localization bounding box prediction, respectively; moreover, 4 denotes the encoding of four coordinates for localization. Furthermore, $M$ is equal to $C\times A$,  where $C$ is the number of classes ($C$ is 80 for the COCO~\cite{lin2014coco} dataset) and $A$ is the number of anchors.

Specifically, with $N_{pos}$ positive samples, our training objective function $L_{rank}$ can be described as the average precision loss $L_{p}$ of each positive sample. The $L_{rank}$ and localization loss $L_{loc}$ on these two sub-nets can be expressed as follows:
\begin{equation}
\label{eq.3}
\begin{small}
\begin{aligned}
L_{rank} =\frac{1}{N_{pos}} \sum_{i^+}\sum_{c^+}\sum_{x^+,y^+} L_{p}(\hat{P}_{i^+,x^+,y^+,c^+}, \hat{\mathbf{P}})
\end{aligned}
\end{small}
\end{equation}
\begin{equation}
\label{eq.4}
\begin{small}
\begin{aligned}
L_{loc} = \frac{1}{N_{pos}} \sum_{i^+} \sum_{x^+,y^+} L_{\text{GIoU}}(\hat{\mathbf{B}}_{i^+,x^+,y^+},\mathbf{B}_{i^+,x^+,y^+})
\end{aligned}
\end{small}
\end{equation}
where $\hat{P}_{i^+,x^+,y^+,c^+} \in R$ and $\hat{\mathbf{P}}$ are the predictions for each sample on a positive location $\{i^+,x^+,y^+,c^+\}$ and all the samples on every location, respectively. We train $M$ binary predictors rather than one multi-class predictor, following~\cite{Lin_2017_retina}.  $\hat{\mathbf{B}}_{i^+,x^+,y^+}\in R^4$ and $\mathbf{B}_{i^+,x^+,y^+}\in R^4$ are the prediction and the ground truth label for localization, respectively, while $\{ i^+,x^+,y^+ \}$ indicates the position of positive samples. As for localization, $L_{\text{GIoU}}$ is the GIoU loss~\cite{Rezatofighi_2018_giou}.

\section {A Pairwise Ranking Perspective on AP Loss}
Note that the predicted probabilities $\hat{\mathbf{P}}$ (before sigmoid function) of all samples are fed into the loss function $L_{p}$. This loss can accordingly exploit the ranking pairs between the positive and negative samples.
We formulate the precision loss as follows:
\begin{equation}
\label{eq.5}
\begin{small}
\begin{aligned}
L_{p}(\hat{P}_{u},\hat{\mathbf{P}}) & = 1 - \text{Precision}(\hat{P}_{u},\hat{\mathbf{P}})=\frac{FP}{TP+FP}\\&=\frac{rank^-(u)}{rank^+(u)+rank^-(u)}
\\&=\frac{\sum\limits_{v \in \mathcal{N}}H(\hat{P}_{v}-\hat{P}_{u})}{\sum\limits_{v \in \mathcal{P}, v\neq u}H(\hat{P}_{v}-\hat{P}_{u})+\sum\limits_{v \in \mathcal{N}}H(\hat{P}_{v}-\hat{P}_{u})} 
\end{aligned}
\end{small}
\end{equation}
where index $u$ is used to replace $\{i^+,x^+,y^+,c^+\}$ for notation convenience, while $rank^+$ and $rank^-$ denote the ranking position of sample $u$ in the positive sample set $\mathcal{P}$ and negative sample set $\mathcal{N}$ respectively. They also indicate the number of True Positives~($TP$) and False Positives~($FP$), respectively. Here, to approximately calculate the ranking position, AP loss adopts the distance function $\mathbf{H}(\cdot)$ between sample pairs, which is a piece-wise step function and can be written as follows:
\begin{equation}
\begin{small}
H(x)=\left\{
\begin{aligned}
0,\qquad  & x<-\delta \\
x/2\delta,\qquad  &  -\delta<=x<=\delta \\
1,\qquad  & \delta<x
\end{aligned}
\right.
\end{small}
\label{eq.6}
\end{equation}
where $\delta$ is a tuning hyper-parameter used to control the slope of $\mathbf{H}(\cdot)$ in $\lbrack-\delta,+\delta\rbrack$.

However, the vector field of prediction errors of $\mathbf{H}(\cdot)$ is not conservative. Thus~\cite{cheng2019aploss} proposed an Error-Driven Update method for AP loss by manually setting the error $g_u$ as the gradient of precision loss $L_p$ \wrt a positive prediction $\hat{P}_u$. Here, $g_u$ is defined as follows:
\begin{equation}
\label{eq.7}
\begin{small}
\begin{aligned}
 &g_u =  \frac{-\sum_{v \in\mathcal{N}}H(\hat{P}_{v}-\hat{P}_{u})}{rank^+(u)+rank^-(u)}
\end{aligned}
\end{small}
\end{equation}
where $g_u$ is used in the same way as $\frac{\partial L_p(\hat{P}_{u},\hat{\mathbf{P}})}{\partial\hat{P}_u}$ in back propagation.
Similarly, $g_v$, the gradients \wrt a negative prediction $\hat{P}_v$ can be formulated as: $g_v=H(\hat{P}_{v}-\hat{P}_{u})/(rank^+(u)+rank^-(u))$,
where $g_v$ is used in the same way as $\frac{\partial L_p(\hat{P}_{u},\hat{\mathbf{P}})}{\partial\hat{P}_v}$ in back propagation.
It is worth noting that $\sum_{v \in \mathcal{N}}g_v = -g_u$; thus, pairwise design can alleviate the imbalance of positive and negative samples.

In fact, the denominator term of the backpropagation gradient $g_u$ shown in Eq.~\eqref{eq.7} is the same as the one of precision loss in Eq.~\eqref{eq.5} and can be considered as a balance constant.
It is evident that optimizing AP loss is equivalent to minimizing the numerator in Eq.~\eqref{eq.5}\cite{oksuz2020alrp}, which can also be considered as the sum of the distance error between each pair of prediction scores, \ie, pairwise error.
Therefore, to reduce the redundancy of analyzing ranking tasks,  we can rewrite the precision loss as pairwise error loss with the unified formalized description following~\cite{burges2005ranknet}.
We here present this unified pairwise error loss $L_{\text{PE}}$ as follows:
\begin{equation}
\begin{small}
\begin{aligned}
L_{\text{PE}}(\hat{P}_{u},\hat{\mathbf{P}}) = -\frac{1}{BC}\sum_{v\in\mathcal{N}}D(\hat{P}_{u}-\hat{P}_{v})
\end{aligned}
\end{small}
\label{eq.8}
\end{equation}
where $BC$ denotes the balance constant, while $\mathbf{D}(\cdot)$ denotes the distance function adopted as the pairwise error.
Finally, we have our ranking loss $L_{rank}$ for all positive samples in a mini-batch: $L_{rank} =\sum_{u} L_{\text{PE}}(\hat{P}_{u}, \hat{\mathbf{P}})/ {N_{pos}}$.

\begin{table}
% \vspace{-0.3cm}
\setlength{\abovecaptionskip}{0.cm}
\setlength{\belowcaptionskip}{-0.cm}
% \makeatletter\def\@captype{table}
\caption{Comparison of distance functions.}
\resizebox{0.45\textwidth}{!}{
\begin{tabular}{c|c|lll}
\toprule[1.5pt]
\multicolumn{1}{c|}{distance function}& \multicolumn{1}{c|}{Error-Driven Update}&
\multicolumn{1}{c}{$\mathrm{AP}$}& \multicolumn{1}{c}{$\mathrm{AP}_{50}$}& \multicolumn{1}{c}{$\mathrm{AP}_{75}$}\\
\hline
\hline
$\mathbf{H}(\cdot)$ & \CheckmarkBold& 37.4& 57.5& 39.2 \\
$\mathbf{S}(\cdot)$ & \CheckmarkBold& 37.3& 57.4& 38.9 \\
$CE(\mathbf{S}(\cdot),0)$ & & 37.3 & 57.2& 39.1 \\
\bottomrule[1.5pt]
\end{tabular}
}
\label{table:comparsion_distant_function}
\vspace{-0.2cm}
\end{table}

\section{What Contributes to Pairwise Error?}
As shown in Eq.~(\ref{eq.8}), the reformulation of the pairwise ranking loss  consists of three essential parts: (1) the distance function $\mathbf{D}(\cdot)$; (2) the balance constant $BC$; (3) the pair selection of $\hat{P}_{v}$ and $\hat{P}_{u}$.
To find out the essential properties of different designs in ranking-based losses, we analyze each part separately and verify the effect of different strategies for pairwise ranking methods~\cite{burges2005ranknet,tan2019l2r,oksuz2020alrp}:

\noindent \textbf{Distance Function:}~\cite{burges2005ranknet,tan2019l2r} adopt sigmoid $\mathbf{S}(\cdot)$ as distance function, which can be described as follows: $S(\hat{P}_{v}-\hat{P}_{u}) = 1/(1+exp(\lambda(\hat{P}_{u}-\hat{P}_{v}))$, where $\lambda$ is a tuning hyper-parameter that controls the slope of the sigmoid function near the origin. By contrast, AP loss uses a non-differentiable function $\mathbf{H}(\cdot)$, described in Eq.~\eqref{eq.6} and employs Error-Driven Update for backpropagation.
However, as shown in the first and second rows of Table~\ref{table:comparsion_distant_function}, when $\mathbf{H}(\cdot)$ in AP loss is replaced with $\mathbf{S}(\cdot)$, the influence on performance is minor.
Additionally, we also use a cross entropy loss $\mathbf{CE}(\cdot,0)$ to replace Error-Driven Update (see appendix for more details).
The results are shown in Table~\ref{table:comparison_balance_constant}, where it can be seen that the usage of Error-Driven Update provides the same performance as CE loss~(\ie, $37.3$).
These results illustrate that the influences on the performance of different distance functions employed in recent pairwise loss~\cite{burges2005ranknet,tan2019l2r,cheng2019aploss} are minor.
% \begin{minipage}[t]{\linewidth}

% \end{minipage}

\noindent \textbf{Balance Constant:}
Consequently, we need a balance constant to normalize the sum of all pairwise errors. 
AP loss uses $rank^+(u)+rank^-(u)$ in Eq.~\eqref{eq.7} as its balance constant.
While~\cite{burges2005ranknet} sets the balance constant as the number of valid negative samples $N_{neg}$, which can keep the loss function simple.
Note that, only if $N_{neg}$ is small enough, we can replace balance constant with $N_{neg}$.
Hence, we use a threshold to define which sample is valid (for more details, please refer to the appendix).
We conduct empirical studies with these two balance constants.
As shown in Table~\ref{table:comparsion_distant_function}, the two different balance constants also yield similar performances~($37.3$ \emph{v.s.} $37.3$).
Although the results suggest that these two balance constant choices are robust, the balance constant design of AP loss is more adaptive and does not require a  valid sample threshold hyper-parameter.
% \begin{minipage}[t]{\linewidth}
\begin{table}
% \vspace{-0.3cm}
\setlength{\abovecaptionskip}{0.cm}
\setlength{\belowcaptionskip}{-0.cm}
% \makeatletter\def\@captype{table}
\caption{Comparison of balance constants.}
\resizebox{\linewidth}{!}{
\begin{tabular}{c|c|lll}
\toprule[1.5pt]
\multicolumn{1}{c|}{Balance Constant}& \multicolumn{1}{c|}{\quad Threshold \quad}& \multicolumn{1}{c}{$\mathrm{AP}$}& \multicolumn{1}{c}{$\mathrm{AP}_{50}$}& \multicolumn{1}{c}{$\mathrm{AP}_{75}$}\\
\hline
\hline
$rank^+(u)+rank^-(u)$ &  & 37.3& 57.4& 38.9\\
$N_{neg}$&\CheckmarkBold  & 37.3& 56.7& 39.4
\\
\bottomrule[1.5pt]
\end{tabular}
}
\label{table:comparison_balance_constant}
% \end{minipage}
\vspace{-0.2cm}
\end{table}

\noindent \textbf{Ranking Pair Selection:} As for the selection of ranking pairs, \ie, $(\hat{P}_{v},\hat{P}_{u})$,
 ALRP loss~\cite{oksuz2020alrp} adopts ATSS~\cite{zhang2020atss} while AP loss adopts an IoU threshold, following~\cite{Lin_2017_retina}.
When we use the ATSS~\cite{zhang2020atss} to select ranking pairs, the performance of pairwise loss is  significantly promoted; specifically, it is boosted from 37.3 to 39.2.
This result may be caused by some important ranking pairs being ignored or many of the selected ranking pairs being inaccurate.
Thus, inappropriate ranking pair selection is a significant barrier to higher accuracy.
% \begin{minipage}{\linewidth}
\begin{table}
% \vspace{+0.3cm}
\makeatletter\def\@captype{table}
\setlength{\abovecaptionskip}{0.cm}
\setlength{\belowcaptionskip}{-0.cm}
    \caption{Varying for sampling strategy on $L_{\text{PE}}$.}
\centering
\resizebox{\linewidth}{!}{
\begin{tabular}{c|lll|lll}
\toprule[1.5pt]
\multicolumn{1}{c|}{Sampling Strategy}& %\multicolumn{1}{c|}{Epoch}&
\multicolumn{1}{c}{$\mathrm{AP}$}& \multicolumn{1}{c}{$\mathrm{AP}_{50}$}& \multicolumn{1}{c|}{$\mathrm{AP}_{75}$}&\multicolumn{1}{c}{$\mathrm{AP_S}$} & \multicolumn{1}{c}{$\mathrm{AP_M}$} & \multicolumn{1}{c}{$\mathrm{AP_L}$}\\
\hline
\hline
IoU threshold& 37.3& 57.4& 38.9& 19.4& 41.7& 51.8 \\
ATSS & 39.2& 59.4& 41.8& 21.4& 43.2& 53.7\\
% PAA  & 39.5& 58.3& 41.5& 20.2& 43.4& 56.5\\
\bottomrule[1.5pt]
\end{tabular}
}
%\vspace{-0.5cm}
\label{tab:M}
% \end{minipage}
\vspace{-0.2cm}
\end{table}

\begin{figure}[!t]
\setlength{\abovecaptionskip}{-0cm}
\setlength{\belowcaptionskip}{-0cm}
  \centering
  \hspace{-15mm}
  \begin{subfigure}{0.44\linewidth}
    \includegraphics[width=\linewidth]{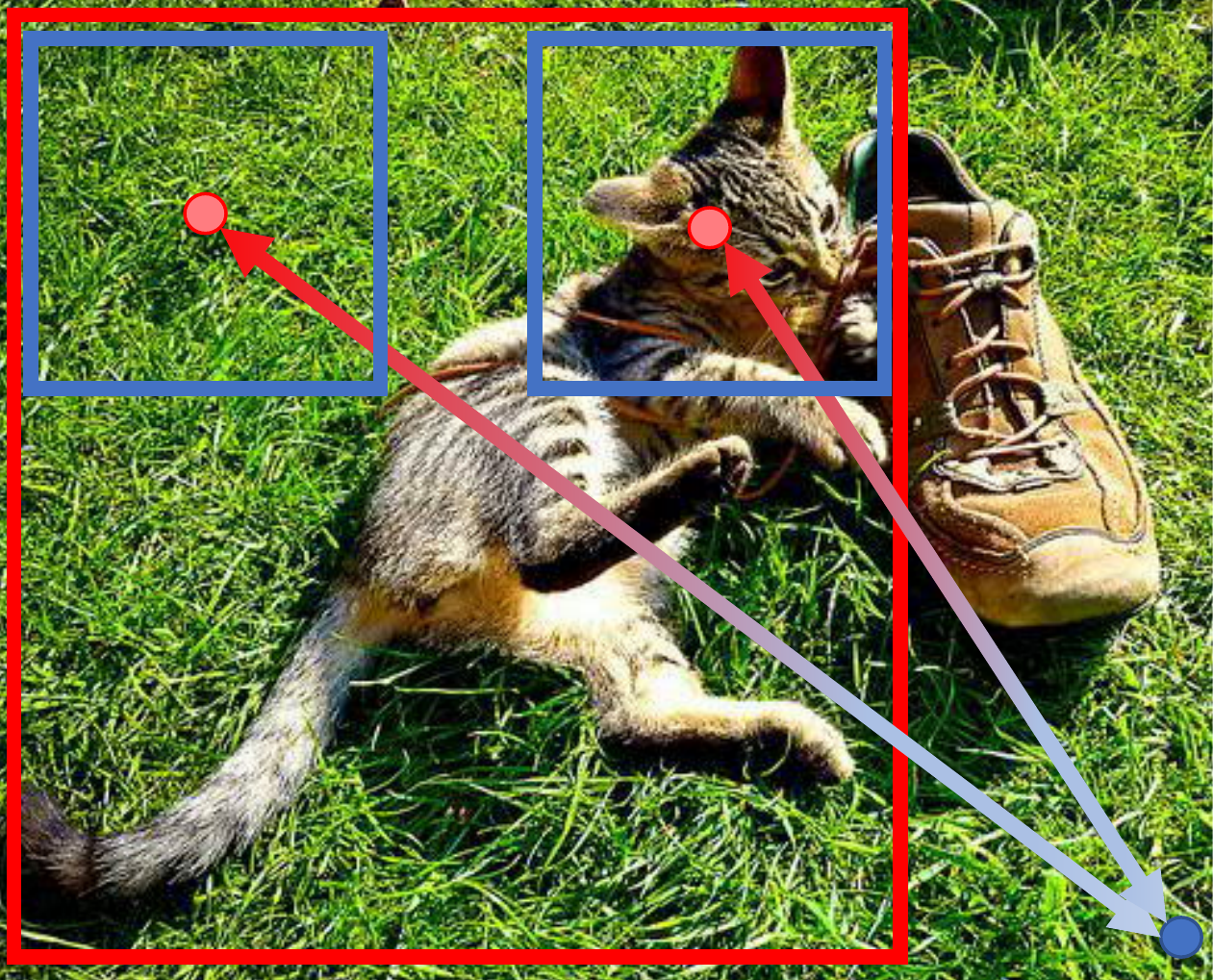}
    % \fbox{\rule{0pt}{2in} \rule{.9\linewidth}{0pt}}
    \caption{}
    \label{fig:short-a}
  \end{subfigure}
  \begin{subfigure}{0.361\linewidth}
  \centering
    \includegraphics[width=\linewidth,angle=90]{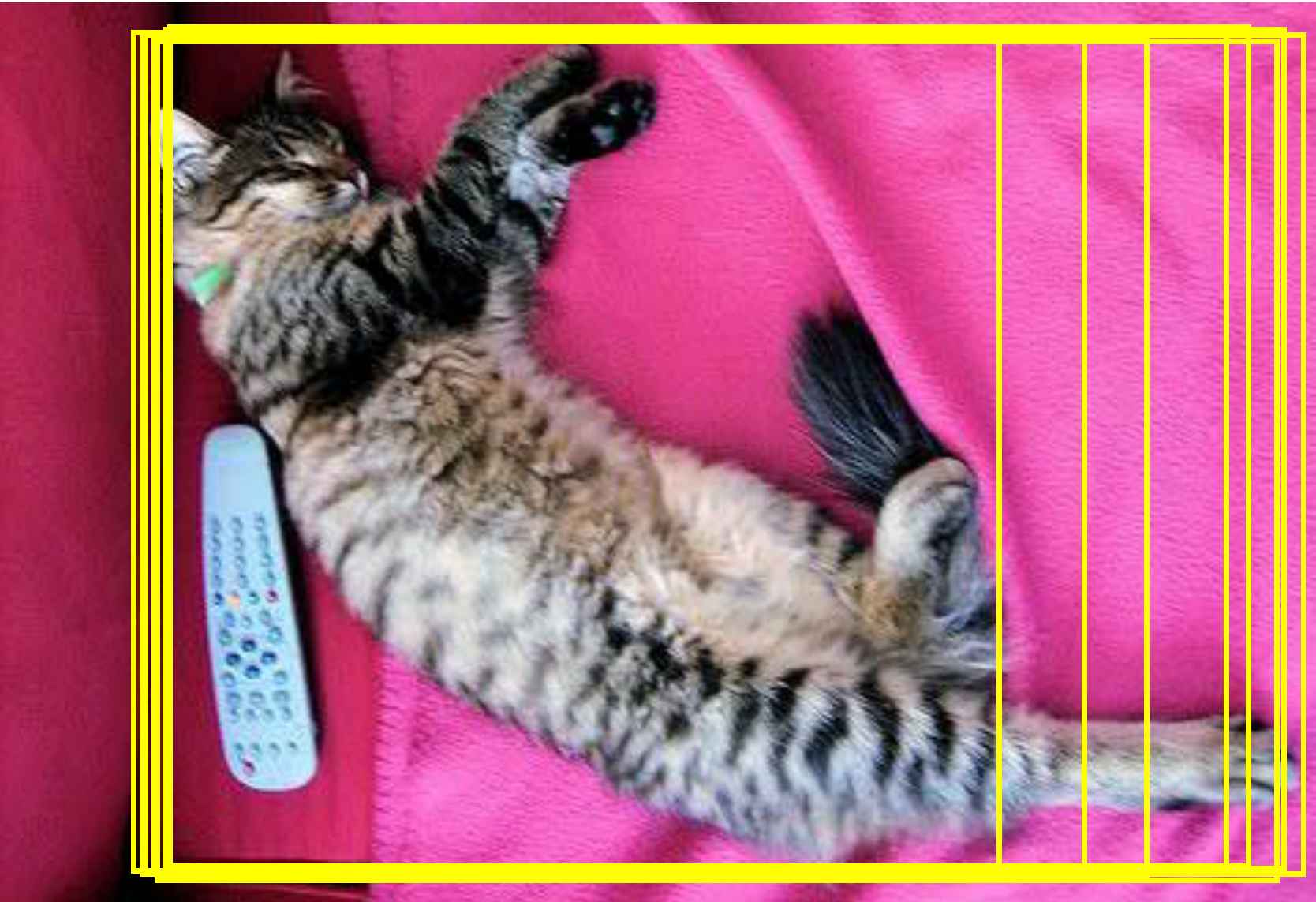}
    % \fbox{\rule{0pt}{2in} \rule{.9\linewidth}{0pt}}
    \caption{}
    \label{fig:short-b}
  \end{subfigure}
  \begin{subfigure}{\linewidth}
    \includegraphics[width=\linewidth]{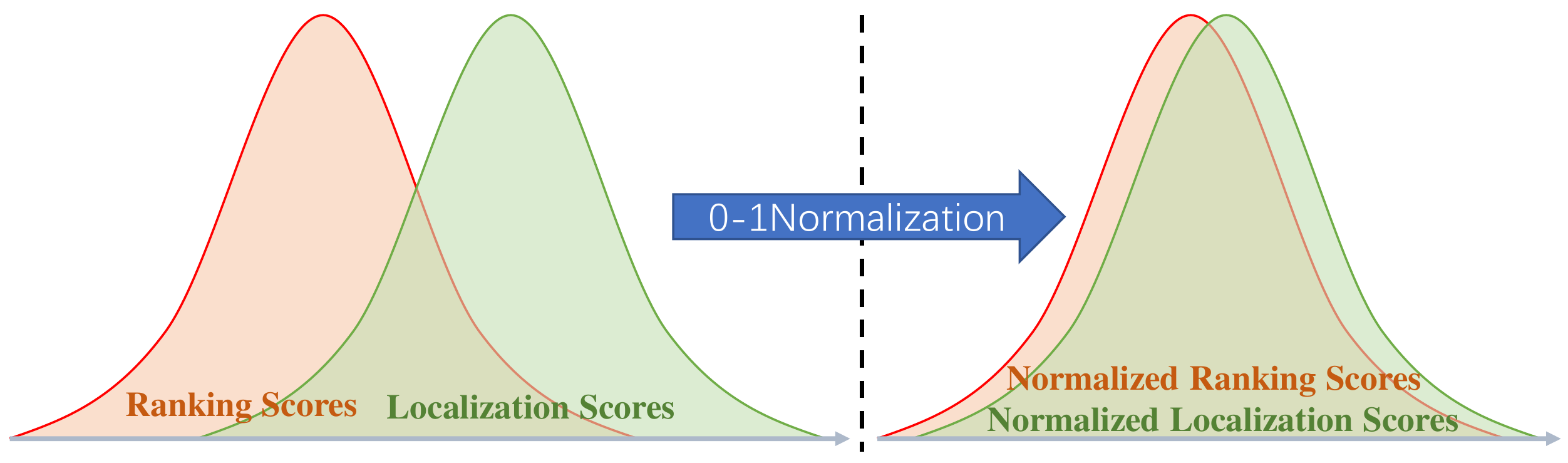}
    % \fbox{\rule{0pt}{2in} \rule{.9\linewidth}{0pt}}
    \caption{}
    \label{fig:short-c}
  \end{subfigure}
    % \subfigure[ ]{\includegraphics[width=0.44\linewidth]{cat_2.pdf}}
    % \subfigure[ ]{\includegraphics[width=0.361\linewidth,angle=90]{pos_cat.pdf}}
    % \subfigure[ ]{\includegraphics[width=\linewidth]{fig_two_dis.pdf}}
  \caption{(a) Different positions in the ground truth box may have the same IoU between their preset anchor box and ground truth. The blue boxes are used to denote anchor boxes, while two-way arrows are used to highlight two ranking pairs. (b) Dense object detection can predict many candidate boxes from positive positions. The box that perfectly predicts this cat is clearly performing better than those that ignore the cat’s foot. (c) \textbf{Left}: The distribution of original ranking scores and localization scores. Although they are in the range of $[0, 1]$, they have different distributions. \textbf{Right}: The distribution of $0$\text{-}$1$ normalized ranking scores and localization scores.}
\vspace{-0.2cm}
\label{fig:cat}
\end{figure}
\noindent \textbf{Analysis of Inappropriate Ranking Pair Selection:}
Current state-of-the-art pair selection strategies~\cite{cheng2019aploss,oksuz2020alrp,qian2020dr} only depend on IoUs between preset anchor boxes and ground truths; thus, there is a lack of exploiting image content.
As can be seen from Fig.~\ref{fig:cat}~(a), the background and foreground may have the same priority, since the distance to the centre of this ground truth box is the same for both of them, leading to an inferior ranking pair selection (\eg, there will be a ranking pair between the two grass positions in Fig.~\ref{fig:cat}~(a)).
Recent studies on adaptive sampling selection~\cite{li2020noisyanchor,kehuang2020mal,kim2020paa} integrate both the predicted category probabilities and localization results into positive sample selection, since the category predictions can represent the image content, while accurate localization results can also achieve better detection accuracy.
To address this barrier, we employ PAA~\cite{kim2020paa} which utilizes both classification loss and localization loss for adaptive sampling selection. 
However, two major problems remain with the existing sampling algorithms when incorporated into the pairwise ranking method:

(1) existing ranking pair selection strategies naturally ignore all important ranking pairs between positive samples~(\ie, $v$ in Eq.~\eqref{eq.8} is limited in the negative sample set $\mathcal{N}$).
As shown in Fig.~\ref{fig:cat}~(b), the detection model may predict many candidate boxes for a specific positive position.
This reveals the existence of localization accuracy disparities between positive samples.
It is accordingly necessary to exploit the localization information for ranking loss. 
If ranking pairs between the positive samples are ignored, the detector is more likely to select unreliable boxes, which has a severely detrimental impact on the final accuracy.

(2) Ranking scores~(\ie, the predicted ranking probabilities $S(\hat{\mathbf{P}})$) used in the sampling processing have different distributions with localization scores~(\ie, IoUs), leading to the attentions of the two tasks being uncoordinated. We illustrate this phenomenon in Fig.~\ref{fig:cat}~(c). The localization scores typically have values close to 1, while the ranking scores generally have relatively lower values.
This is because the training stops before the ranking scores of positive samples go to $1$.
In particular, when the $\hat{P}_{u}$ of positive sample is relatively higher than $\hat{P}_{v}$, the distance function $\mathbf{D}(\cdot)$ will drop to $0$.
To this end, PAA~\cite{kim2020paa} will focus more on image content than localization accuracy, because classification loss will obtain a higher value than localization loss at the end of the training.

\setlength{\textfloatsep}{0.5cm}
\begin{algorithm}[t]
    \small
    \SetAlgoLined
    \LinesNumbered
    \KwIn{ \\
           \parbox{\dimexpr\textwidth-2\algomargin\relax}{$\mathcal{P}$ is a set of positive positions}
           \parbox{\dimexpr\textwidth-2\algomargin\relax}{$\mathcal{N}$ is a set of negative positions}
           {$\hat{\mathbf{I}}$ are the IoUs between boxes $\hat{\mathbf{B}}$ predicted at the corresponding position and ground truth}				
         }
    \KwOut{ \\
    \parbox{\dimexpr\textwidth-2\algomargin\relax}{$\mathcal{A}$ is a set of adaptive negative position sets}
    }
    $\mathcal{A} \leftarrow \varnothing$\;
    \ForAll{$u \in \mathcal{P}$}{
        build an empty negative sample set for each positive sample position $u$: $\mathcal{A}_u \leftarrow \varnothing$\;
    %   calculate the IoU between ground truth and prediction on position $u$: $I_u= \text{IoU} (\hat{\mathbf{B}}_u,\mathbf{B}_u)$\;
       \ForAll{$v \in \mathcal{P}, v\neq u$}{
    %   calculate the IoU between ground truth and prediction on position $u$: $I_u= \text{IoU} (\hat{\mathbf{B}}_v,\mathbf{B}_v)$\;
    
        \If{$\hat{I}_v < \hat{I}_u$}{{If its IoU is smaller than that of $p$, this sample should be considered an adaptive False Positive (aFP) sample in original $\mathcal{P}$ set $\mathcal{A}_u = \mathcal{A}_u \cup v$}\;
            }
       }
    $\mathcal{A}_u=\mathcal{A}_u \cup \mathcal{N}$\;
    $\mathcal{A}=\mathcal{A} \cup \{\mathcal{A}_u\}$\;
    }
    return $\mathcal{A}$\;
   
\caption{Adaptive Ranking Pair Selection}
\label{alg:1}
\end{algorithm}

\section{Adaptive Ranking Pair Selection}

We propose ARPS to calculate Adaptive Pairwise Error between positive samples.
Our method focuses on dynamically selecting negative samples $\mathcal{A}_u$ in positive set $\mathcal{P}$ according to their localization qualities.
Notably, each positive sample $u$ is assigned with a different set $\mathcal{A}_u$ to replace $\mathcal{N}$ in Eq.~\eqref{eq.8}.
Here, APE loss $L_{\text{APE}}$ can be written as follows:%$L_{\text{APE}}(\hat{P}_{u},\hat{\mathbf{P}}) = -\sum_{v\in\mathcal{A}_u}D(\hat{P}_{v}-\hat{P}_{u})/BC$.
\begin{equation}
\begin{small}
\begin{aligned}
L_{\text{APE}}(\hat{P}_{u},\hat{\mathbf{P}}) = -\frac{1}{BC}\sum_{v\in\mathcal{A}_u}D(\hat{P}_{u}-\hat{P}_{v})
\end{aligned}
\end{small}
\label{eq.9}
\end{equation}
where the balance constant $BC$ is set as $rank^+(u)+rank^-(u)$, {while the} distance function is set as cross entropy with sigmoid $CE(\mathbf{S}(\cdot),0)$. 
In a departure from previous works~\cite{Jiang_2018_iounet,tian2019fcos,xu2020asso,li2020generalized}, adaptive pairwise error integrates localization quality-related information into extra ranking pairs between positive samples rather than regressing a certain value (e.g., center-ness in~\cite{tian2019fcos} and IoU in~\cite{li2020generalized}).
Our method is adaptive and does not bring any hyper-parameters, which eases training overloads for different applications.
With the dynamic expansion of the negative sample set, our proposed APE loss considers the ranking pairs in the positive samples and further boosts the detection accuracy.
%s 
%Moreover, our method

\setlength{\algomargin}{1em}
\SetAlCapNameFnt{\footnotesize}
\SetAlCapFnt{\footnotesize}

\noindent \textbf{Description:} Alg.~\ref{alg:1} describes how the proposed method works for an existing negative sample set and dynamically expands it.
For each positive position $u\in \mathcal{P}$ with IoU $\hat{I}_u$, we add the samples $v$ with lower IoU $\hat{I}_v$ than $\hat{I}_u$ into a dynamically expanded negative sample set $\mathcal{A}_u$ as shown in Alg.~\ref{alg:1}.
After we store all additional negative samples from $\mathcal{P}$, we merge $\mathcal{A}_u$ with $\mathcal{N}$ in lines 9 to 10.
After we {employ} the ARPS to calculate adaptive pairwise error in Eq.~(\ref{eq.9}), it would be interesting to explore the obstacle part in the original AP metric.
Hence, we re-write APE loss in AP formula and analyze which parts of AP loss are modified by ARPS.
\begin{equation}
\begin{small}
\begin{aligned}
L_{\text{APE}}&(\hat{P}_{u},\hat{\mathbf{P}}) = -\frac{1}{BC}\sum_{v\in\mathcal{A}_u}D(\hat{P}_{u}-\hat{P}_{v})\\
&=\frac{\sum\limits_{v \in \mathcal{N}}D(\hat{P}_{v}-\hat{P}_{u})+\sum\limits_{v \in \mathcal{P}, I_v<I_u}D(\hat{P}_{v}-\hat{P}_{u})}{\sum\limits_{v \in \mathcal{P}, v\neq u}D(\hat{P}_{v}-\hat{P}_{u})+\sum\limits_{v \in \mathcal{N}}D(\hat{P}_{v}-\hat{P}_{u})}\\
& =\frac{FP+aFP}{(TP-aFP)+(FP+aFP)}\\
\end{aligned}
\end{small}
\label{eq.10}
\end{equation}
where $BC$ is also set as $rank^+(u)+rank^-(u)$; moreover, a sample should be considered as an adaptive False Positive (aFP) sample if its IoU is smaller than that of $p$. 

In conclusion, {the question of how} to define TPs and FPs is an important one for dense object detection, as not all positive samples are the best detection results.
ARPS {accordingly} identifies some adaptive False Positive (aFP) samples in the original TP set, and can thus be considered {to provide} a more accurate AP metric.

\begin{figure}
%\vspace{-1.2cm}
\setlength{\abovecaptionskip}{0.cm}
\setlength{\belowcaptionskip}{-0cm}
\centering
    \includegraphics[width =0.49\textwidth]{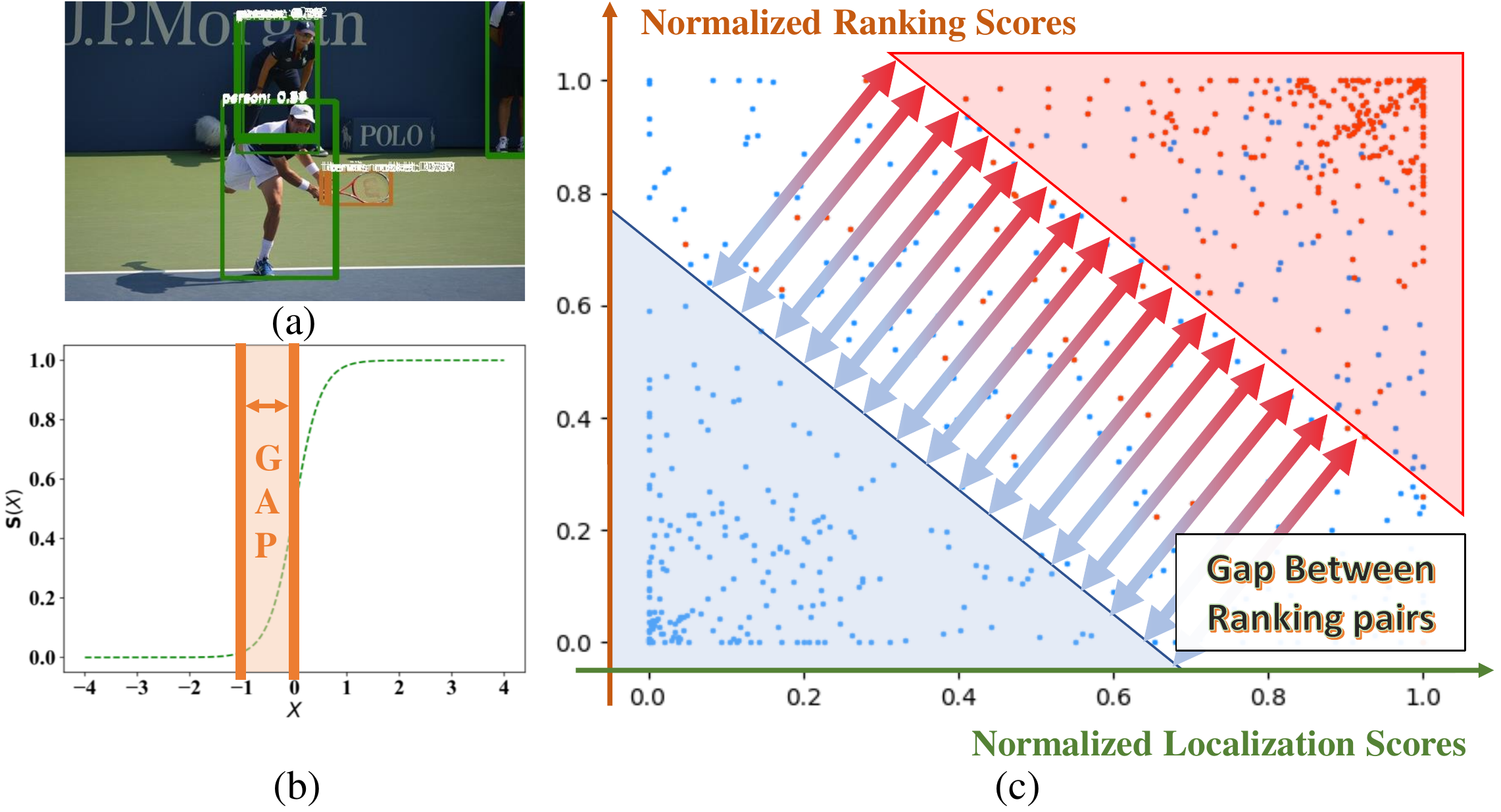}
    \caption{Illustration of gaps between ranking pairs. (a) The massive candidate box before NMS can have a high IoU with the ground truth, while all remaining negative samples have 0 IoU. (b) $\mathbf{S}(\cdot)$ and its design, which lead to a small gap between ranking scores. (c) Ideal gap between ranking pairs in the view of normalized ranking and localization scores.}
\vspace{-0.2cm}
\label{fig:gap}
\end{figure}

Additionally, to reduce the inaccurate ranking pairs in ARPS {that are} selected by the preset anchor’s IoU, we focus on image contents by investigating ranking scores and localization scores. 
Considering there are massive numbers of candidate boxes before the NMS algorithm, positive samples around objects can {receive comparatively} higher localization scores while the IoUs of negative samples are normally close to $0$.
This means that a large localization score gap exists between positive and negative samples. 
Instead, it can be inferred from Fig.~\ref{fig:gap}~(b) that the ranking scores of the positive and negative {samples} only have a small gap between them (\eg, the value of the distance function will be close to $0$ when $\hat{P}_{u}-\hat{P}_{v}$ goes to $1$, which means that after sigmoid function, their gap $S(\hat{P}_{u})-S(\hat{P}_{v})$ is {far} smaller than $1$).

We here employ PAA \cite{kim2020paa} to exploit {the} information of these pair gaps. Specifically, we first use 0-1 normalization for each instance to reduce the scale differences between the ranking score and localization score.
We can then observe from Fig.~\ref{fig:gap}~(c) that samples naturally gather close to two ends~(\ie, $(0,0)$ and $(1,1)$), and the blank area between the blue and red shades is an ideal separation that we expected.
However, there are still many outliers of the two clusters, \ie, the small blue and red dots in the gap.
If the ranking pairs are formulated as every different {potential} combination of two samples from the two clusters~(the bottom-left and top-right region in Fig.~\ref{fig:gap}~(c)),
the pairwise error will in turn enlarge this gap and control the outliers as the training {progresses}.
Therefore, we convert the ranking pair selection problem into a two-cluster discrimination problem on the two normalized scores.
Following PAA~\cite{kim2020paa}, Gaussian Mixture Model~(GMM) is used to construct positive and negative clusters.

\section{Experiments}

\subsection{Dataset and Implementation Details}
\label{sec:7.1}
Experiments are conducted on {the} MS-COCO 2017 dataset~\cite{lin2014coco}. We use a split set \texttt{train-2017~(115k images)} for training and validation set \texttt{val-2017~(5k images)} for the ablation study and comparison with other losses. 
\texttt{test-dev} split is used to  test advanced additions.
The COCO-style average precision~(AP) is employed as the performance metric. The implementation is based on Pytorch 1.4, CUDA 10 and mmdetection~\cite{mmdetection}.
Unless {otherwise} specified, all hyper-parameters are set to default, as in RetinaNet~\cite{Lin_2017_retina} and FCOS~\cite{tian2019fcos}.
The ranking score threshold of NMS is 0.15 and IoU threshold of NMS is 0.6.
The batch size is $16$ and $4$ GPUs are used for training~(we use 4 $\times$ GTX 2080Ti-11G for training on ResNet-50~\cite{He_2016_resnet} and 4 $\times$ RTX Titan-24G for training with advanced methods, \ie, advanced backbone and multi-scale training). SGD is used as an optimizer and warmed up for the first 500 iterations.

\noindent \textbf{RetinaNet 512:} Here, we follow the baseline, \ie, AP loss, and employ SSD-style~\cite{liu2016ssd} data augmentations (including \texttt{Photo Metric Distortion} and \texttt{Min IoU Random Crop}~\cite{mmdetection}) for comparable evaluations.
Moreover, to improve this baseline, we employ GIoU~\cite{Rezatofighi_2018_giou} loss to obtain more accurate localization results.
Here, the max pair number $Q$ (see appendix) here is $1,000,000$.
The resolution of the input images is $512\times512$ and the initial learning rate is $0.004$.
For $48$-epochs training, we divide the learning rate by $10$ at $32$ epochs and again at 44 epochs. %For 96 epochs training, we divide learning rate by 10 at 64 and again at 88 epochs.

\noindent \textbf{FCOS 800:} We here omit the center-ness branch and make the pipeline more elegant.
To {ensure} focus on more important samples, we also use category prediction to weight the GIoU loss following~\cite{li2020generalized}, and use IoU to weight APE loss following~\cite{wu2019iou}. The weight of ranking loss is set to $1.0$.
The max pair number $Q$ here is $100,000$~(see appendix). The resolution of the input images is $800\times1333$ with the same ratio of raw images. The initial learning rate is $0.01$. For $24$-epochs training, we divide the learning rate by $10$ at $18$ and again at $22$ epochs.

\subsection{Ablation Study}

\begin{table}[t]%{\linewidth}
% \makeatletter\def\@captype{table}
% \vspace{-0.3cm}
\setlength{\abovecaptionskip}{0.cm}
\setlength{\belowcaptionskip}{0.cm}
% \makeatletter\def\@captype{table}
\caption{Overall contributions of {the} proposed methods over baseline. $\Diamond$ denotes the usage of a higher learning rate. * denotes replacing the clustering input of PAA~\cite{kim2020paa} with normalized ranking and localization scores.}
\resizebox{\linewidth}{!}{
\begin{tabular}{c|c|lll}%|lll}
\toprule[1.5pt]
\multicolumn{1}{c|}{Ranking Loss} &\multicolumn{1}{c|}{Sampling Strategy} & \multicolumn{1}{c}{$\mathrm{AP}$}& \multicolumn{1}{c}{$\mathrm{AP}_{50}$}& \multicolumn{1}{c}{$\mathrm{AP}_{75}$}\\ %\multicolumn{1}{c}{$\mathrm{AP_S}$} & \multicolumn{1}{c}{$\mathrm{AP_M}$} & \multicolumn{1}{c}{$\mathrm{AP_L}$}\\
\hline
\hline
% AP Loss &IoU threshold &35.4& 58.1& 37.0\\
AP Loss$\Diamond$ & IoU threshold &37.3 & 57.4& 38.9\\
Adaptive Pairwise Error& IoU threshold &38.3 &55.8 &41.1 \\
Adaptive Pairwise Error& ATSS  &39.9 &58.3 &42.6 \\
Adaptive Pairwise Error& PAA* & 41.1& 59.6& 43.7\\
\bottomrule[1.5pt]

\end{tabular}}
\label{table:ablation}
\vspace{-0.2cm}
\end{table}
\begin{table}
%\vspace{-0.35cm}
\setlength{\abovecaptionskip}{0.cm}
\setlength{\belowcaptionskip}{0.cm}
\caption{Contribution of APE loss {in} association with localization scores.}
\resizebox{\linewidth}{!}{
\begin{tabular}{c|lll}
\toprule[1.5pt]
\multicolumn{1}{c|}{Method}& \multicolumn{1}{c}{$\mathrm{PCC}$}& \multicolumn{1}{c}{$\mathrm{SCC}$}& \multicolumn{1}{c}{$\mathrm{KCC}$}\\% \multicolumn{1}{c}{$\mathrm{AP_S}$} & \multicolumn{1}{c}{$\mathrm{AP_M}$} & \multicolumn{1}{c}{$\mathrm{AP_L}$}\\
\hline
\hline
Pairwise Error & 0.3742& 0.3790& 0.2538 \\
Adaptive Pairwise Error & 0.4940& 0.5034& 0.3449\\ 
\bottomrule[1.5pt]
\end{tabular}
}
\label{table:correlation}
\vspace{-0.2cm}
\end{table}

We conduct ablation experiments on ResNet-50-FPN~\cite{He_2016_resnet,Lin_2017_fpn} to evaluate the effect of each component of our proposed method.
To begin with, we implement our approach on RetinaNet to evaluate individual components relative to the baseline method, \ie, AP loss~\cite{cheng2019aploss}.
However, RetinaNet is an obsolete detector with lower performance than the currently most popular method, \ie, FCOS.
Therefore, we also test our method on the state-of-the-art detector FCOS~(ATSS) to illustrate the superiority of APE loss over recent classification losses.

\noindent \textbf{Effect of Adaptive Pairwise Error:} We conduct experiments with our implementation of Pairwise Error on RetinaNet, where $\lambda$ is set to $8$ (see appendix for the varying of $\lambda$).
As shown in Table~\ref{table:ablation}, by replacing AP loss with APE loss, we can improve the AP and $\mathrm{AP}_{75}$ by $1.0$ and $2.2$, respectively. This demonstrates that our proposed APE loss improves detection accuracy.
Moreover, to better understand the reasons why APE loss is effective, we adopt IoUs and ranking scores of detection results with higher localization scores~(\ie, IoUs larger than 0.5) as two lists of data and calculate their Pearson Correlation Coefficient~(PCC), Spearman Correlation Coefficient~(SCC) and Kendall rank Correlation Coefficient~(KCC).
The results are {presented} in Table~\ref{table:correlation}.
From the table, it can be seen that APE loss promotes PCC by $0.12$, SCC by $0.12$ and KCC by $0.09$.
As quantitative measurements of correlation, the three coefficients are promoted significantly, indicating that APE loss can associate the category predictions with localization qualities and {thereby} boost NMS accuracy.

\noindent \textbf{Effect of Normalized Scores Input of PAA:}
To study the necessity of two normalized scores for PAA, we test the original PAA~\cite{kim2020paa} that adopts Focal loss~\cite{Lin_2017_retina} and GIoU loss~\cite{Rezatofighi_2018_giou} for clustering.
As the results in Table~\ref{table:comparison_clustering_input} show, when we replace normalized ranking and classification scores with Focal loss and GIoU loss, the performance drops from $41.1$ to $39.4$.
We contend that it is because the scales of ranking and localization gap in these losses are not well aligned; localization loss would be smaller, and its information could be easily ignored.
However, PAA works well in classification methods owing to the fact that these two scores are trained to be equal~\cite{li2020generalized,zhang2020vfl} and could have similar distribution scales.
We additionally test different clustering input features; normalized ranking scores and localization scores independently achieve $38.9$ and $33.9$ AP, illustrating the association of two scores brings $2.2$ AP performance gain.
Moreover, as shown in Table~\ref{table:comparison_clustering_input}, when normalization processing on the two scores is omitted, the performance also drops to $40.5$, illustrating the normalization brings a $0.6$ AP performance gain.

\begin{table}[t!]
% \vspace{-0.3cm}
\setlength{\abovecaptionskip}{0.cm}
\setlength{\belowcaptionskip}{0.cm}
\makeatletter\def\@captype{table}
\caption{Comparison of different clustering input scores of PAA. Here all scores are 0-1 normalized for each instance.}
\resizebox{\linewidth}{!}{
\begin{tabular}{c|c|lll}%|lll}
\toprule[1.5pt]
\multicolumn{1}{c|}{GMM Clustering Input}& \multicolumn{1}{c|}{0-1 Normalization}& \multicolumn{1}{c}{$\mathrm{AP}$}& \multicolumn{1}{c}{$\mathrm{AP}_{50}$}& \multicolumn{1}{c}{$\mathrm{AP}_{75}$}\\%& \multicolumn{1}{c}{$\mathrm{AP_S}$} & \multicolumn{1}{c}{$\mathrm{AP_M}$} & \multicolumn{1}{c}{$\mathrm{AP_L}$}\\
\hline
\hline

% Localization Scores &  & 33.1& 52.7& 33.5\\%& 16.2& 36.8& 50.8 \\
% Ranking Scores &  & 38.9& 57.6& 41.3\\%& 21.3& 42.6& 55.5\\ 
Focal Loss + GIoU Loss&   &39.4 &58.1 &41.4\\% &20.7 &43.3 & 56.0\\
Ranking + Localization Scores &  & 40.5 &57.6 & 43,5\\
Ranking + Localization Scores & \CheckmarkBold & 41.1& 59.6& 43.7\\%& 22.9& 45.2& 57.3\\
\bottomrule[1.5pt]
\end{tabular}
}
\label{table:comparison_clustering_input}
\vspace{-0.2cm}
\end{table}
\begin{table}[t!]
\setlength{\abovecaptionskip}{0.cm}
\setlength{\belowcaptionskip}{0.cm}
\makeatletter\def\@captype{table}
\caption{Comparison of different detectors with our method.}
\resizebox{\linewidth}{!}{
\begin{tabular}{c|lll|lll}
\toprule[1.5pt]
\multicolumn{1}{c|}{Detector}& \multicolumn{1}{c}{$\mathrm{AP}$}& \multicolumn{1}{c}{$\mathrm{AP}_{50}$}& \multicolumn{1}{c|}{$\mathrm{AP}_{75}$}& \multicolumn{1}{c}{$\mathrm{AP_S}$} & \multicolumn{1}{c}{$\mathrm{AP_M}$} & \multicolumn{1}{c}{$\mathrm{AP_L}$}\\
\hline
\hline
RetinaNet(512) & 41.1& 59.6& 43.7& 22.9& 45.2& 57.3\\ 
FCOS(800)   & 41.5& 59.2& 44.7& 23.6& 46.1& 54.8 \\
\bottomrule[1.5pt]
\end{tabular}
}
\label{table:comparison_detector}
\vspace{-0.2cm}
\end{table}

\begin{table}[t!]
\setlength{\abovecaptionskip}{0.cm}
\setlength{\belowcaptionskip}{0.cm}
\makeatletter\def\@captype{table}
\caption{Training time comparison of different training conditions.}
\resizebox{\linewidth}{!}{
\begin{tabular}{c|c|c|c}
\toprule[1.5pt]
\multicolumn{1}{c|}{Ranking Loss}& \multicolumn{1}{c|}{Resolution}& \multicolumn{1}{c|}{Learning Rate}&  \multicolumn{1}{c}{Epoch} \\
\hline
\hline
 AP Loss & 512 & 0.001 &100\\
 AP Loss & 512 & 0.004 &48 \\
 Adaptive Pairwise Error & 512 & 0.004 &48 \\
 Adaptive Pairwise Error & 800 & 0.01  &24\\
\bottomrule[1.5pt]
\end{tabular}
}
\label{table:traing_time}
\vspace{-0.2cm}
\end{table}

\begin{table*}[!ht]
%\vspace{-0.1cm}
\setlength{\abovecaptionskip}{0.cm}
\setlength{\belowcaptionskip}{-0.cm}
\newcommand{\tabincell}[2]{\begin{tabular}{@{}#1@{}}#2\end{tabular}}
\setlength{\belowdisplayskip}{1cm}
\centering
\caption { Here we omit center-ness branch~\cite{tian2019fcos} or IoU branch~\cite{kim2020paa} in our method. Data Aug. denotes the SSD-style~\cite{liu2016ssd} data augmentation. $^\dagger$ denotes multi-scale training, $^\ddagger$ denotes multi-scale testing.}
\resizebox{\textwidth}{!}{
\begin{tabular}{l|c|c|c|c|lll|lll}
%\toprule
\toprule[1.5pt]
    \multicolumn{1}{c|}{Loss Function}         &
    \multicolumn{1}{c|}{Method}&
    %\multicolumn{1}{c|}{Additional Branch} &
    \multicolumn{1}{c|}{Data Aug.} &
    \multicolumn{1}{c|}{Epoch} &
    \multicolumn{1}{c|}{Backbone}         & \multicolumn{1}{c}{$\mathrm{AP}$} & \multicolumn{1}{c}{$\mathrm{AP}_{50}$} & \multicolumn{1}{c|}{$\mathrm{AP}_{75}$} & %\multicolumn{1}{c|}{$\mathrm{AP}_{90}$} &
    \multicolumn{1}{c}{$\mathrm{AP_S}$} & \multicolumn{1}{c}{$\mathrm{AP_M}$} & \multicolumn{1}{c}{$\mathrm{AP_L}$}\\ %\multicolumn{1}{c}{$\mathrm{AR_1}$} & \multicolumn{1}{c}{$\mathrm{AR_{10}}$} & \multicolumn{1}{c}{$\mathrm{AR_{100}}$}
\hline
\hline
% Baseline (CE)                           & 33.8   & 52.8   & 35.4  & \textbf{19.6}  & 38.7    & 44.2  \\
\tabincell{l}{$Classification\ Loss$~(on \texttt{val-2017}):\\ }&&&&&&&&\\
\tabincell{l}{Focal Loss~\cite{Lin_2017_retina}~(on \texttt{test-dev}) \\ } & Retina($500$)  & \XSolidBrush  & 12 &ResNet-50-FPN     & 32.5 &50.9 &34.8 &13.9 &35.8 &46.7\\
% \tabincell{l}{Focal Loss~\cite{Lin_2017_retina} \\ } & Retina($800$)  & \XSolidBrush  & 12 &ResNet-50-FPN     & 35.6 &55.6 &38.2 &19.1 &39.2 &46.3\\
\tabincell{l}{GHM Loss~\cite{li2019ghm}\\ } & Retina($800$)  & \XSolidBrush  & 12 &ResNet-50-FPN     & 35.8 &55.5 &38.1 &19.6 &39.6 &46.7\\
\tabincell{l}{Focal Loss~\cite{zhang2020atss} \\ } & ATSS($800$)  & \XSolidBrush  & 12 &ResNet-50-FPN     & 39.2 & 56.5 &40.7 &20.6 & 42.1 &49.1\\
\tabincell{l}{Quality Focal Loss~\cite{li2020generalized}} & ATSS($800$) & \XSolidBrush & 12 &ResNet-50-FPN     & 39.9   & 58.5   & 43.0   & 22.4  & 43.9    & 52.7 \\
\tabincell{l}{Varifocal Loss~\cite{zhang2020vfl}} & ATSS($800$) & \XSolidBrush & 12 &ResNet-50-FPN     & 40.2   & 58.2   & 44.0   & 23.9  & 44.1    & 52.2 \\
\tabincell{l}{Focal Loss~\cite{kim2020paa}) \\ } & PAA($800$)  & \XSolidBrush  & 18 &ResNet-50-FPN     & 41.1 &59.4 &44.3 &23.5 &45.4 &54.3\\
\tabincell{l}{$Ranking\ Loss$~(on \texttt{val-2017}):\\ }&&&&&&&&\\
\tabincell{l}{Average Precision Loss~\cite{cheng2019aploss,oksuz2020alrp}} & Retina($512$) & \CheckmarkBold & 100 &ResNet-50-FPN     & 35.4   & 58.1   & 37.0   & 19.2  & 39.7    & 49.2 \\
\tabincell{l}{Distributional Ranking Loss~\cite{qian2020dr}} & Retina($800$) & \XSolidBrush & 12 &ResNet-50-FPN     & 37.4   & 56.0   & 40.0   & 20.8  & 41.2    & 50.5 \\
\tabincell{l}{aLRP Loss~\cite{oksuz2020alrp}} & ATSS($512$) & \CheckmarkBold & 100 &ResNet-50-FPN    &41.0 &61.2 &43.3 &22.9 &44.7 &55.8\\
\tabincell{l}{RankDetNet Losses~\cite{Liu_2021_rankDetNet}} & ATSS($800$) &\XSolidBrush & 24 &ResNet-50-FPN   &40.7 &58.4 &44.0 &23.8 &44.8 &52.8\\
\tabincell{l}{Rank \& Sort Loss~\cite{Oksuz_2021_ranksort}} & PAA($800$) &\XSolidBrush & 12 &ResNet-50-FPN   &41.0 &59.1 &44.5 &23.9 &44.9 &53.5\\
\tabincell{l}{$Ours$~(on \texttt{val-2017}):\\ }&&&&&&&&\\
\tabincell{l}{Adaptive Pairwise Error} & Retina($512$)  & \CheckmarkBold  & 48 &ResNet-50-FPN  & $38.3$  &$55.8$ &$41.1$ &$21.0$ &$42.5$ &$53.1$   \\
\tabincell{l}{Adaptive Pairwise Error} & ATSS($512$)  & \CheckmarkBold  & 48 &ResNet-50-FPN  & $39.9$  &  $58.3$  & $42.6$  & $21.9$     & $44.1$     & $55.0$  \\
\tabincell{l}{Adaptive Pairwise Error} & ATSS($800$)  & \XSolidBrush  & 24 &ResNet-50-FPN  & $40.8$  &$58.4$ &$44.0$ &$23.0$ &$44.5$ &$54.5$   \\
\tabincell{l}{Adaptive Pairwise Error} & PAA*($512$)  & \CheckmarkBold  & 48 &ResNet-50-FPN  & $41.1$  &  $59.6$  & $43.7$  & $22.9$     & $45.2$     & $57.3$  \\
\tabincell{l}{Adaptive Pairwise Error} & PAA*($512$)  & \CheckmarkBold  & 96 &ResNet-50-FPN  & $42.9$  &  $61.3$  & $46.1$  & $25.0$     & $47.1$     & 59.1  \\
\tabincell{l}{Adaptive Pairwise Error} & PAA*($800$)  & \XSolidBrush  & 24 &ResNet-50-FPN  & $41.5$& $59.2$& $44.7$& $23.6$& $46.1$& $54.8$\\

\hline
\tabincell{l}{$Losses \ with \ Adv.\ Tricks$ (on \texttt{test-dev}):  \\ }&&&&&&&&\\
\tabincell{l}{Average Precision Loss$^\ddagger$~\cite{cheng2019aploss}} & Retina($512$) & \CheckmarkBold & 100 &ResNet-101-FPN     & 42.1   & 63.5   & 46.4   & 25.6  & 45.0    & 53.9 \\
\tabincell{l}{Distributional Ranking Loss$^\dagger{}^\ddagger$}~\cite{qian2020dr}& Retina($800$) & \XSolidBrush & 18 &ResNeXt-101-FPN     & 44.7   & 63.8   & 48.7  & 28.2  & 47.4    & 56.2 \\
\tabincell{l}{Focal Loss$^\dagger$~\cite{zhang2020atss} \\ } & ATSS($800$)  & \XSolidBrush  & 24 &ResNeXt-101-FPN-DCN     & 47.7 & 66.5 &51.9 & 29.7& 50.8 &59.4\\
\tabincell{l}{Generalize Focal Loss$^\dagger$~\cite{li2020generalized} \\ } & ATSS($800$)  & \XSolidBrush  & 24 &ResNeXt-101-FPN-DCN     & 48.2 & 67.4 &52.6 & 29.2& 51.7 &60.2\\
\tabincell{l}{$Ours \ with \ Adv.\ Tricks$ (on \texttt{test-dev}):\\ }&&&&&&&&\\
\tabincell{l}{Adaptive Pairwise Error$^\dagger$} & PAA*($800$)  &
\XSolidBrush  & 24 &ResNeXt-101-FPN-DCN  & 49.0  &67.1 &53.0 &29.9 &52.9 & 63.4\\
\tabincell{l}{Adaptive Pairwise Error$^\dagger$} & PAA*($800$)  & \XSolidBrush  & 36 &Res2Net-101-FPN-DCN  & $\mathbf{50.0}$   & $\mathbf{67.7}$ & $\mathbf{53.9}$ & $\mathbf{31.8}$ &  $\mathbf{55.3}$ & $\mathbf{66.7}$  \\
\bottomrule[1.5pt]
\end{tabular}}
\label{table:sota}
\vspace{-0.2cm}
\end{table*}

\noindent \textbf{Effect of Higher Resolution and Learning Rate:}
To facilitate comparative evaluation of our method with state-of-the-art losses, we incorporate it into the most popular detector with the higher input resolution, \ie, FCOS(800).
As the results in Table~\ref{table:comparison_detector} show, training on FCOS(800) can improve our method by $0.4$ AP.
Note that we here discard SSD-style augmentation.
We argue that this promotion occurs due to the higher input resolution, since the main performance gap between RetinaNet and FCOS is addressed by ATSS~\cite{zhang2020atss}.
We also record the comparison of training epochs as shown in Table.~\ref{table:traing_time}.
It can be observed that our method only needs a quarter of the AP loss training epochs, \ie, $24$ rather than $100$.
However, due to the slow convergence speed of baseline AP loss, we still train APE loss with more epochs than recent losses\cite{Lin_2017_retina,kim2020paa,li2020generalized,zhang2020vfl}.
If a higher learning rate~(\ie, 0.04) is used, we can also train AP loss within $48$ epochs, the same as ours.
Thus, we argue that the promotion of training efficiency occurs largely because of the higher learning rate and higher image resolution than the baseline.

\subsection{Comparison with State-of-the-Art Losses}
Finally, we compare APE loss with state-of-the-art loss functions on val-2017, as shown in Table~\ref{table:sota}.
Here, every method only uses single-model and single-scale testing, omits 1.5 stage regression (as in~\cite{H_2020_borderdet,zhang2020vfl}) and utilizes ResNet-50-FPN~\cite{He_2016_resnet,Lin_2017_fpn}.
APE loss achieves $41.5$ on the FCOS~\cite{tian2019fcos} detector without center-ness branch.
We also verify our method on ATSS, PAA and RetinaNet.
The performances of APE loss are superior to that of recent state-of-the-art losses.
Furthermore, we test our method with advanced additions on test-dev, \ie, multi-scale training and stronger backbones (ResNext~\cite{Xie_2017_resnext}, Res2Net~\cite{gao2021res2net}) with Deformable ConvNets v2~\cite{Zhu_2019_dcnv2}.
As shown in  Table~\ref{table:sota}, those additions boost our methods to $50.0$ AP.

\section{Conclusion}
In this paper, we revisit the AP loss from a pairwise ranking perspective for dense object detection. In the process, we reveal an essential fact that proper ranking pair selection plays an important role in producing accurate detection results compared with the distance function and balance constant. Therefore, we propose a novel strategy, Adaptive Ranking Pair Selection~(ARPS), by providing more complete and accurate ranking pairs. We first exploit the localization information into extra rank pairs with the Adaptive Pairwise Error, which can also be considered as a more accurate form of AP loss. We then use normalized ranking scores and localization scores to split the positive and negative samples. The proposed method is very simple and achieves performance comparable to existing classification and ranking methods.

% %%%%%%%%% REFERENCES
\section*{Acknowledgement}
This work is supported by the Major Project for New Generation of AI under Grant No.~2018AAA0100400, the National Natural Science Foundation of China (Grant No.~62176047) and Beijing Natural Science Foundation (Z190023).

\newpage

{\small
\bibliographystyle{ieee_fullname}
\bibliography{egbib}
}

\end{document}

\title{-Appendix- \\ Revisiting AP Loss for Dense Object Detection: Adaptive Ranking Pair Selection}

\maketitle

\section{Distant Function Selection}
% \noindent \textbf{Sigmoid Function or Piece-wise Step Function:}
We evaluate the distant function with piece-wise step function $\mathbf{H}(\cdot)$ and sigmoid function $\mathbf{S}(\cdot)$, as shown in Fig.~\ref{fig:h(x)} and Fig.~\ref{fig:s(x)} respectively.
The experimental results based on RetinaNet~\cite{Lin_2017_retina} are given in Table~\ref{table:varying_lambda_delta}. We can observe that the performance gap between piece-wise step function $\mathbf{H}(\cdot)$ and sigmoid function $\mathbf{S}(\cdot)$ is only $0.1\%$ in terms of AP~($37.4$ \emph{v.s.} $37.3$).
The results demonstrate that these two distance functions have no essential difference.
%Moreover, the hyper-parameter $\lambda$ in $\mathbf{S}(\cdot)$ is not sensitive.
In this paper, we use $\lambda = 8$ for all experiments.
\begin{minipage}[t!]{\textwidth}
\vspace{0.3cm}
\begin{minipage}[t!]{0.24\textwidth}
% \vspace{-0.3cm}
% \setlength{\abovecaptionskip}{0.cm}
% \setlength{\belowcaptionskip}{0.cm}
\makeatletter\def\@captype{figure}
\centering
\includegraphics[width =\textwidth]{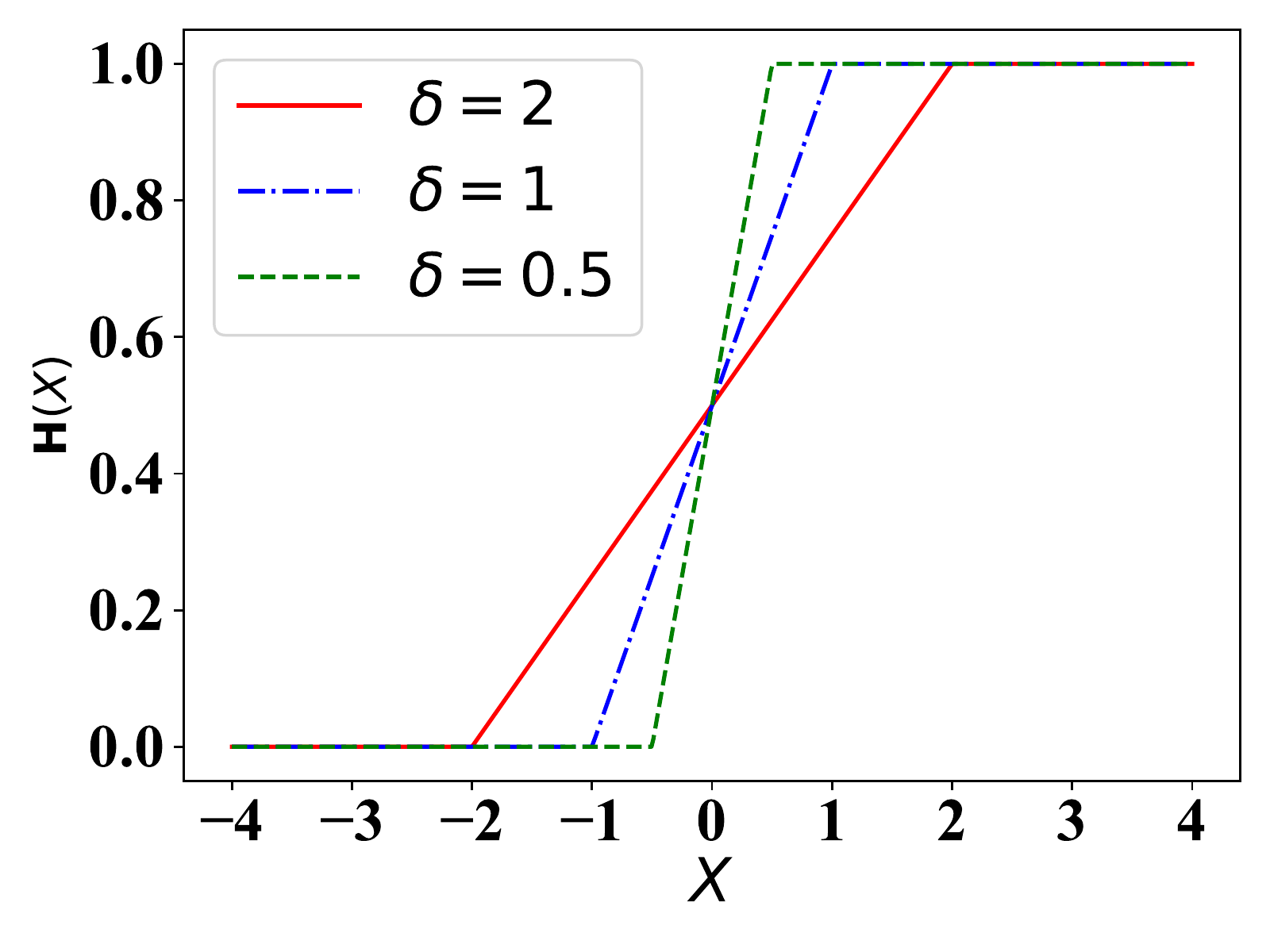}
\caption{$\mathbf{H}(\cdot)$.}
\label{fig:h(x)}
\end{minipage}
\begin{minipage}[t!]{0.24\textwidth}
% \vspace{0.3cm}
% \setlength{\abovecaptionskip}{0.cm}
% \setlength{\belowcaptionskip}{0.cm}
\makeatletter\def\@captype{figure}
 \centering
 \includegraphics[width =\textwidth]{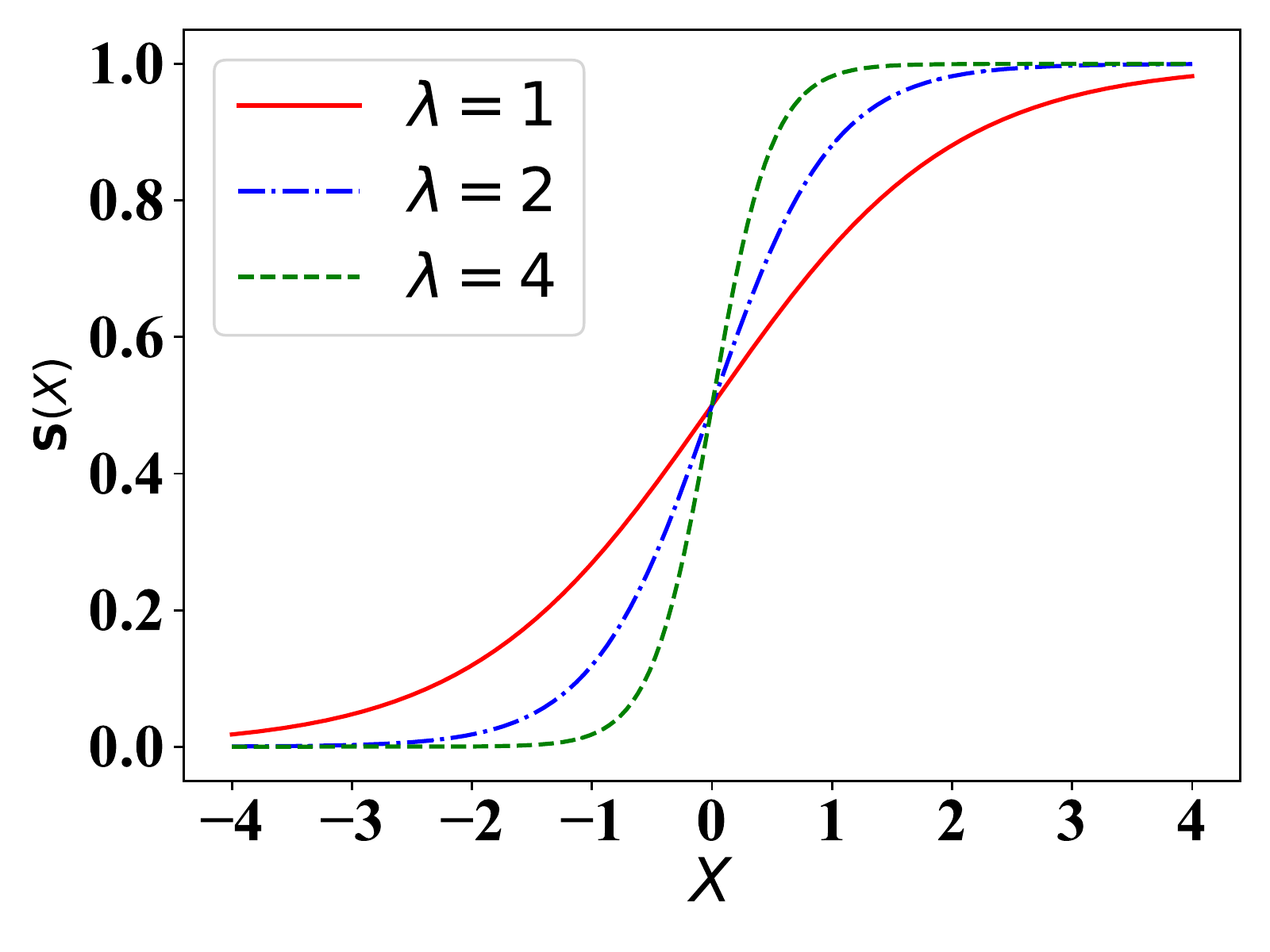}
\caption{$\mathbf{S}(\cdot)$.}
    % \vspace{-0.3cm}
\label{fig:s(x)}
\end{minipage}
\end{minipage}

\section{The Equivalence between Cross Entropy and Error-Driven Update}
Here we find that if pairwise error loss has the same gradients form as Eq.(7) in the main paper, then Error-Driven Update can be omitted for simplicity.
To keep the numerator of pairwise error gradients as the same as Eq.(7) in the main paper, we follow the common practice on cross entropy loss which adds a logistic function to sigmoid function.
To start with, $\mathbf{S}(\cdot)$ is replaced with $\mathbf{CE}(\mathbf{S}(\cdot),0)/\lambda$, which can be written as:
\begin{equation}
\begin{small}
\begin{aligned}
&\frac{1}{\lambda }CE(S(\hat{P}_{v}-\hat{P}_{u}),0)  \\
 &=-\frac{1}{\lambda }((1-0)\cdot log(1-S(\hat{P}_{v}-\hat{P}_{u}))+0\cdot (S(\hat{P}_{v}-\hat{P}_{u})))\\
 &=-\frac{1}{\lambda }log(1-S(\hat{P}_{v}-\hat{P}_{u}))\\
\end{aligned}
\end{small}
\label{eq.10}
\end{equation}
where the gradient of this distance function w.r.t $S(\hat{P}_{v}-\hat{P}_{u})$ can be calculated as:
\begin{equation}
\begin{small}
\begin{aligned}
\frac{\partial CE(S(\hat{P}_{v}-\hat{P}_{u}),0)}{\lambda \partial S(\hat{P}_{v}-\hat{P}_{u})} = \frac{1}{\lambda (1-S(\hat{P}_{v}-\hat{P}_{u}))}
\end{aligned}
\end{small}
\label{eq.11}
\end{equation}
Since the gradient of $S(\hat{P}_{v}-\hat{P}_{u})$ w.r.t $\hat{P}_{u}$ can be written as:
\begin{equation}
\small
\label{eq.12}
\frac{\partial S(\hat{P}_{v}-\hat{P}_{u})}{\partial\hat{P}_{u}} = -\lambda S(\hat{P}_{v}-\hat{P}_{u})(1-S(\hat{P}_{v}-\hat{P}_{u}))
\end{equation}
Therefore, we can have the the gradient of distance function w.r.t $\hat{P}_{u}$:
\begin{equation}
\small
\label{eq.13}
\begin{aligned}
&\frac{\partial CE(S(\hat{P}_{v}-\hat{P}_{u}),0)}{\lambda \partial \hat{P}_{u}} = \frac{\partial CE(S(\hat{P}_{v}-\hat{P}_{u}),0)}{
\lambda \partial S(\hat{P}_{v}-\hat{P}_{u})} \cdot \frac{\partial S(\hat{P}_{v}-\hat{P}_{u})}{\partial\hat{P}_{u}}\\
&= \frac{1}{\lambda (1-S(\hat{P}_{v}-\hat{P}_{u}))} \cdot (-\lambda S(\hat{P}_{v}-\hat{P}_{u})(1-S(\hat{P}_{v}-\hat{P}_{u})))\\
&=-S(\hat{P}_{v}-\hat{P}_{u})
\end{aligned}
\end{equation}
Also, to keep the denominator term $BC$ as the same as Eq.~(7) in the main paper, we detach it from backpropagation and treat it as a constant.
Note that, after employing these two tricks (\ie, cross entropy and detaching), we can have the same gradient of our pairwise error (\ie, $(-\sum_{v \in\mathcal{N}}S(\hat{P}_{v}-\hat{P}_{u}))/(rank^+(u)+rank^-(u))$) as AP loss, which theoretically leads to similar performances. The experimental results in Table~1 in the main paper also demonstrate that.

\begin{table}[t!]
% \setlength{\abovecaptionskip}{0.cm}
% \setlength{\belowcaptionskip}{0.cm}
% \makeatletter\def\@captype{table}
% \vspace{-0.5cm}
\caption{Varying delta and lambda for distance function.}
\resizebox{\linewidth}{!}{
\begin{tabular}{c|lll||c|lll}
% \toprule[1.5pt]
\hline
\hline
\multicolumn{1}{c|}{$\delta$}& \multicolumn{1}{c}{$\mathrm{AP}$}& \multicolumn{1}{c}{$\mathrm{AP}_{50}$}& \multicolumn{1}{c||}{$\mathrm{AP}_{75}$}&
 \multicolumn{1}{c|}{$\lambda$}&\multicolumn{1}{c}{$\mathrm{AP}$} & \multicolumn{1}{c}{$\mathrm{AP}_{50}$} & \multicolumn{1}{c}{$\mathrm{AP}_{75}$}\\
\hline
\hline
1 & 37.0 & 57.6& 39.2& 2& 36.4& 57.1& 37.9\\
% \hline
% % \tabincell{l}{$$\\ }&&&&&&&&\\
% Ours (RetinaNet):&&&&\\
0.5 & 37.4 & 57.5& 39.2& 4& 36.9& 57.5& 38.7\\
0.25 & 36.8& 56.3& 38.7& 8& 37.3& 57.4& 38.9\\
0.125 & 35.1& 53.8& 36.6& 16& 36.5& 55.9& 38.3\\
% \bottomrule[1.5pt]
\hline
\hline
\end{tabular}
}
\label{table:varying_lambda_delta}
\end{table}

\section{Threshold for Selecting Valid Negative Samples}
% \noindent \textbf{Threshold for Selecting Valid Negative Sample:}
In training processing, the number of negative samples $N_{neg}$ is enormous and might overwhelm the loss.
To solve this issue, we utilize a larger margin threshold $T$ to filter out easy negative samples, as shown in Fig.~\ref{fig:easy_pairs}.
Specifically, we set a valid indicator for each pairwise error to ignore easy pairs. Here we describe indicator function $\mathbbm{1}_{uv}$ as:
\begin{equation}
\begin{small}
\mathbbm{1}_{uv}=\left\{
\begin{aligned}
1,\qquad  & \hat{P}_{v}- \hat{P}_{u}>T \\
0,\qquad  & else
\end{aligned}
\right.
\end{small}
\end{equation}
Then $N_{neg}$ is formulated as: ${N_{neg}} = \sum_{v \in \mathcal{N}}\mathbbm{1}_{uv}$. We also study the impact of different thresholds on detection accuracy. As shown in Table~\ref{table:varying_th_bc}, when $T=0.25$, $N_{neg}$ provides the same performance as $rank^+(u)+rank^-(u)$. This demonstrates the selection of these two balance constants is robust.

\begin{minipage}[t!]{\textwidth}
\begin{minipage}[t!]{0.5\textwidth}
\setlength{\abovecaptionskip}{-0cm}
\setlength{\belowcaptionskip}{0.cm}
\makeatletter\def\@captype{figure}
% \caption{The comparison of different clustering input scores of ARPS.}
\centering
\includegraphics[width=0.75\textwidth]{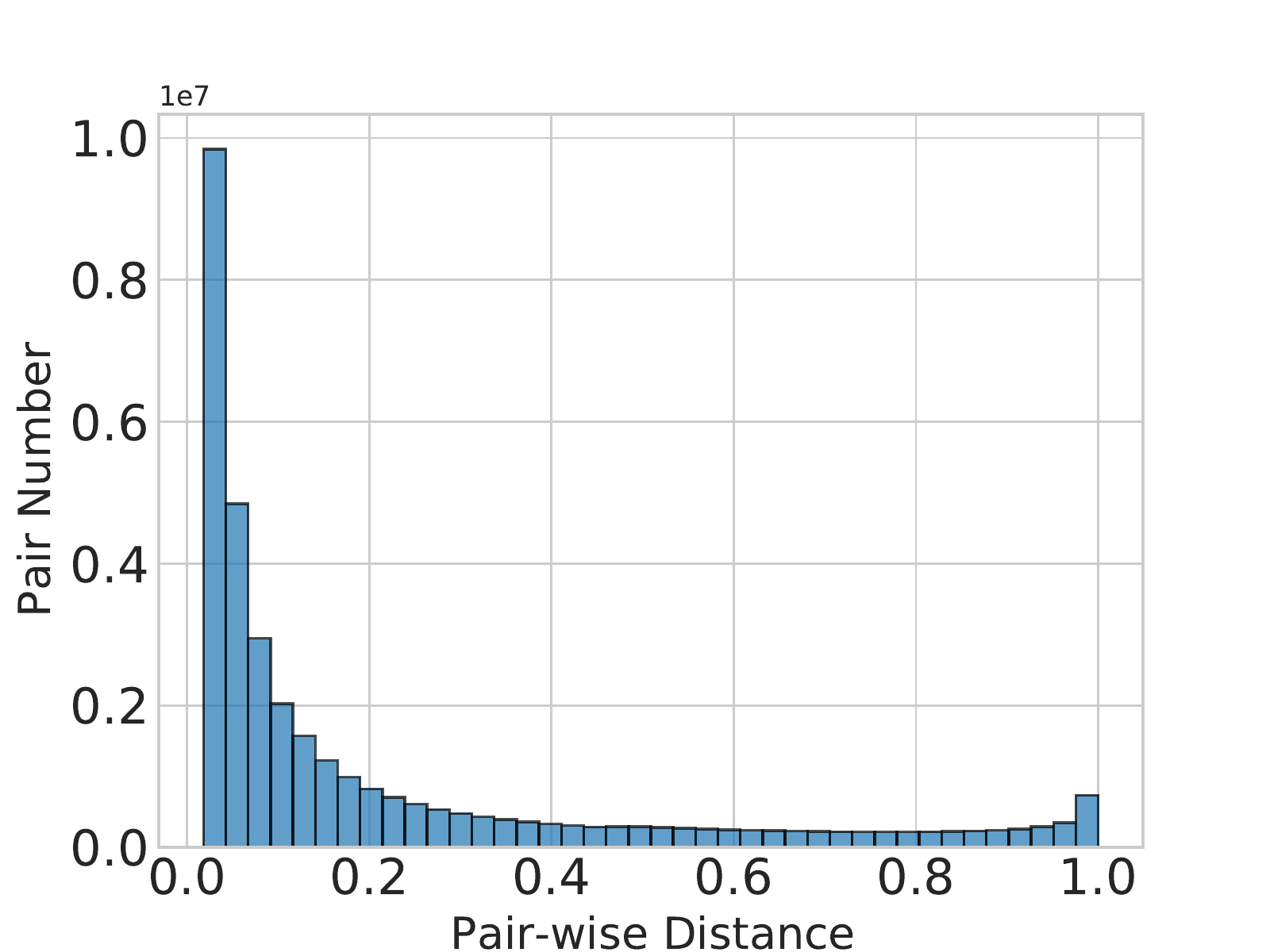}
\label{fig:easy_pairs}
\caption{The comparison with Hard Pair Mining}
\end{minipage}
\end{minipage}

\section{Maximum Pair Number}
In our experiments, the memory (11GB) of \texttt{2080TI} GPU can be ran out because of the extreme large number of pair $\begin{Bmatrix}\hat{P}_{v}, \hat{P}_{u}\end{Bmatrix}$. Thus we adopt a simple yet efficient trick; constricting the input number of pairs.

Here we denote the maximum input number of pairs by $Q$~(\ie the maximum length of $\mathcal{A}_u$ for $L_{\text{APE}}$). Specifically, we manually choose the top $Q$ predictions $\hat{P}_{v}$ of negative samples in $\mathcal{A}_u$. We conduct experiments varying $Q$ for APE loss on FCOS, and the results are shown in Table.~\ref{tab:M}.
When $Q$ is greater than $100,000$, the performance will no longer be improved.
It can be concluded from the results that the promotion from large $Q$ becomes minor as the gradually increasing of $Q$.
\begin{table}[t!]
\setlength{\abovecaptionskip}{0.cm}
\setlength{\belowcaptionskip}{0.cm}
% \makeatletter\def\@captype{table}
\caption{Varying $th$ for $N_{neg}$.}
\resizebox{\linewidth}{!}{
  \begin{tabular}{c|c|lll}%l|lll}
\toprule[1.5pt]
\multicolumn{1}{c|}{Balance Constant}& \multicolumn{1}{c|}{$T$} & \multicolumn{1}{c}{$\mathrm{AP}$}& \multicolumn{1}{c}{$\mathrm{AP}_{50}$} &\multicolumn{1}{c}{$\mathrm{AP}_{75}$}\\%& \multicolumn{1}{c}{$\mathrm{AP_S}$} & \multicolumn{1}{c}{$\mathrm{AP_M}$} & \multicolumn{1}{c}{$\mathrm{AP_L}$}\\
\hline
\hline
$rank^+(u)+rank^-(u)$& N/A & 37.3& 57.4 & 38.9 \\
\hline
$N_{neg}$ &0 &36.8  & 57.1& 38.8\\
% $N_{neg}(th = 0.1)$ & \\ 
$N_{neg}$ & 0.2 & 37.2 & 57.0& 38.9\\ 
$N_{neg}$ &0.25 &37.3& 56.7 & 39.4 \\
$N_{neg}$ &0.3 &36.9 & 56.3 & 38.7\\
% $N_{neg}(th = 0.4)$ & 3\\
$N_{neg}$ &0.5 & 35.2 &53.3& 37.1\\
\bottomrule[1.5pt]
\end{tabular}
}
\label{table:varying_th_bc}
\end{table}

\begin{table}
\vspace{-0.3cm}
\setlength{\abovecaptionskip}{0.cm}
\setlength{\belowcaptionskip}{-0.cm}
    \caption{Varying for $Q$ on FCOS~\cite{tian2019fcos}.}
\centering
% \resizebox{\linewidth}{!}{
\begin{tabular}{l|lll}%|lll}
\toprule[1.5pt]
\multicolumn{1}{c|}{$Q$}& %\multicolumn{1}{c|}{Epoch}&
\multicolumn{1}{c}{$\mathrm{AP}$}& \multicolumn{1}{c}{$\mathrm{AP}_{50}$}& \multicolumn{1}{c}{$\mathrm{AP}_{75}$}\\ %\multicolumn{1}{c}{$\mathrm{AP_S}$} & \multicolumn{1}{c}{$\mathrm{AP_M}$} & \multicolumn{1}{c}{$\mathrm{AP_L}$}\\
\hline
\hline
10,000& 37.6& 54.3& 40.0\\
50,000 & 39.7& 57.3& 42.3\\
100,000& 40.0& 58.1& 42.4\\
200,000& 40.0& 58.1& 42.6\\
\bottomrule[1.5pt]
\end{tabular}
% }
\label{tab:M}
\vspace{-0.5cm}
\end{table}

{\small
\bibliographystyle{ieee_fullname}
\bibliography{egbib}
}